\documentclass[letterpaper, 10 pt, conference]{ieeeconf}  

\IEEEoverridecommandlockouts                              

\usepackage{cite}
\usepackage[nolist]{acronym}
\usepackage{amsmath}
\usepackage{amssymb}
\usepackage{graphicx}
\usepackage{subcaption}
\usepackage{xcolor}
\usepackage{makecell}
\usepackage{placeins} 
\usepackage{balance}
\graphicspath{{pic/}}
\DeclareGraphicsExtensions{.pdf,.png,.jpg}
\usepackage{amsmath,amssymb}
\usepackage{amsmath,amssymb,amsfonts,mathtools}
\usepackage{float}
\title{\LARGE \bf
ReSeFlow: \underline{Re}ctifying \underline{SE}(3)-Equivariant Policy Learning \underline{Flow}s
}

\author{Zhitao Wang$^{1,\dag}$ and Yanke Wang$^{2,\dag}$ and Jiangtao Wen$^{3}$, \IEEEmembership{Fellow, IEEE} and \\ Roberto Horowitz$^{4}$, \IEEEmembership{Senior Member, IEEE} and Yuxing Han$^{1, *}$, \IEEEmembership{Senior Member, IEEE}
\thanks{$\dag$ Equal contribution.}
\thanks{* Corresponding authors: Yuxing Han.}
\thanks{*This work was supported by Shenzhen Startup Funding (No. QD2023014C) supported by Meituan}
\thanks{$^{1}$Zhitao Wang and Yuxing Han are with Tsinghua University, Shenzhen International Graduate School, Shenzhen, China {\tt\small wzt22@mails.tsinghua.edu.cn, yuxinghan@sz.tsinghua.edu.cn}}
\thanks{$^{2}$Yanke Wang is with the Division of Business and Hospitality Management (BHM), College of Professional and Continuing Education (CPCE), The Hong Kong Polytechnic University (PolyU), Hong Kong SAR, China  {\tt\small yanke.wang@cpce-polyu.edu.hk}}
\thanks{Jiangtao Wen is with NYU Shanghai, China {\tt\small  jw9263@nyu.edu}}
\thanks{Roberto Horowitz is with Department of Mechanical Engineering, University of California, Berkeley, USA {\tt\small  horowitz@berkeley.edu}}%
\thanks{This work was submitted to 2026 IEEE International Conference on Robotics \& Automation.}
}

\begin{acronym}
\acrodef{reseflow}[ReSeFlow]{Rectifying \ensuremath{SE(3)}-Equivariant Policy Learning Flows}
\acrodef{dp}[DP]{Diffusion Policy}
\acrodef{ros}[ROS]{Robot Operating System}
\acrodef{se3}[\ensuremath{SE(3)}]{Special Euclidean group}
\acrodef{edf}[EDFs]{Equivariant Descriptor Fields}
\acrodef{ode}[ODE]{Ordinary Differential Equation}
\acrodef{rf}[RF]{rectified flow}
\acrodef{etseed}[ET-SEED]{Efficient Trajectory-Level \ensuremath{SE(3)}-Equivariant Diffusion Policy}
\end{acronym}

\begin{document}

\maketitle
\thispagestyle{empty}
\pagestyle{empty}

\begin{abstract}
Robotic manipulation in unstructured environments requires the generation of robust and long-horizon trajectory-level policy with conditions of perceptual observations and benefits from the advantages of $SE(3)$-equivariant diffusion models that are data-efficient. However, these models suffer from the inference time costs. Inspired by the inference efficiency of rectified flows, we introduce the rectification to the $SE(3)$-diffusion models and propose the ReSeFlow, i.e., Rectifying $SE(3)$-Equivariant Policy Learning Flows, providing fast, geodesic-consistent, least-computational policy generation. Crucially, both components employ $SE(3)$-equivariant networks to preserve rotational and translational symmetry, enabling robust generalization under rigid-body motions. With the verification on the simulated benchmarks, we find that the proposed ReSeFlow with only one inference step can achieve better performance with lower geodesic distance than the baseline methods, achieving up to a 48.5\% error reduction on the painting task and a 21.9\% reduction on the rotating triangle task compared to the baseline's 100-step inference. This method takes advantages of both $SE(3)$ equivariance and rectified flow and puts it forward for the real-world application of generative policy learning models with the data and inference efficiency.
\end{abstract}


\section{Introduction}
\label{sec:intro}

The visual and point-cloud observations from complex environment are increasingly providing the robots with possibility of capturing, understanding the world, and formulating robust motions \cite{han2025neupan,hao2025learn}. Especially, robotic manipulation is greatly benefiting from generative models conditioned with vision models in 2D and 3D cases for generating desirable trajectory-level action policies \cite{zhang2025autoregressive,hao2025disco,liu2025diff9d}. This task faces challenges owing to limitation of generalization of data-driven models, computational costs of iterative generative models, and unsatisfactory accuracy of the generating process. Although recent imitation learning approaches and generative models can produce plausible trajectories \cite{su2025motion,wen2025tinyvla,kang2025robotic}, researchers are still struggling with the difficulties caused by non-Euclidean geometry of robotic actions \cite{yu2025efficient}, long-horizon policy inference \cite{chen2025deformpam}, and multi-modal functionality of the models in manipulation \cite{wen2025tinyvla}. In particular, owing to the nature of the robotic actions in \ac{se3}, straightforward application of Euclidean models to data often leads to fragile policies that fail to generalize to complex and challenging manipulation-involved scenarios with divergent object configurations, viewpoints, or action uncertainties, and data- and inference-efficient learning procedure remains a persistent challenge.
\setlength{\textfloatsep}{6pt plus 0pt minus 2pt}
\begin{figure}[!t]
    \centering
    \includegraphics[width=0.4\textwidth]{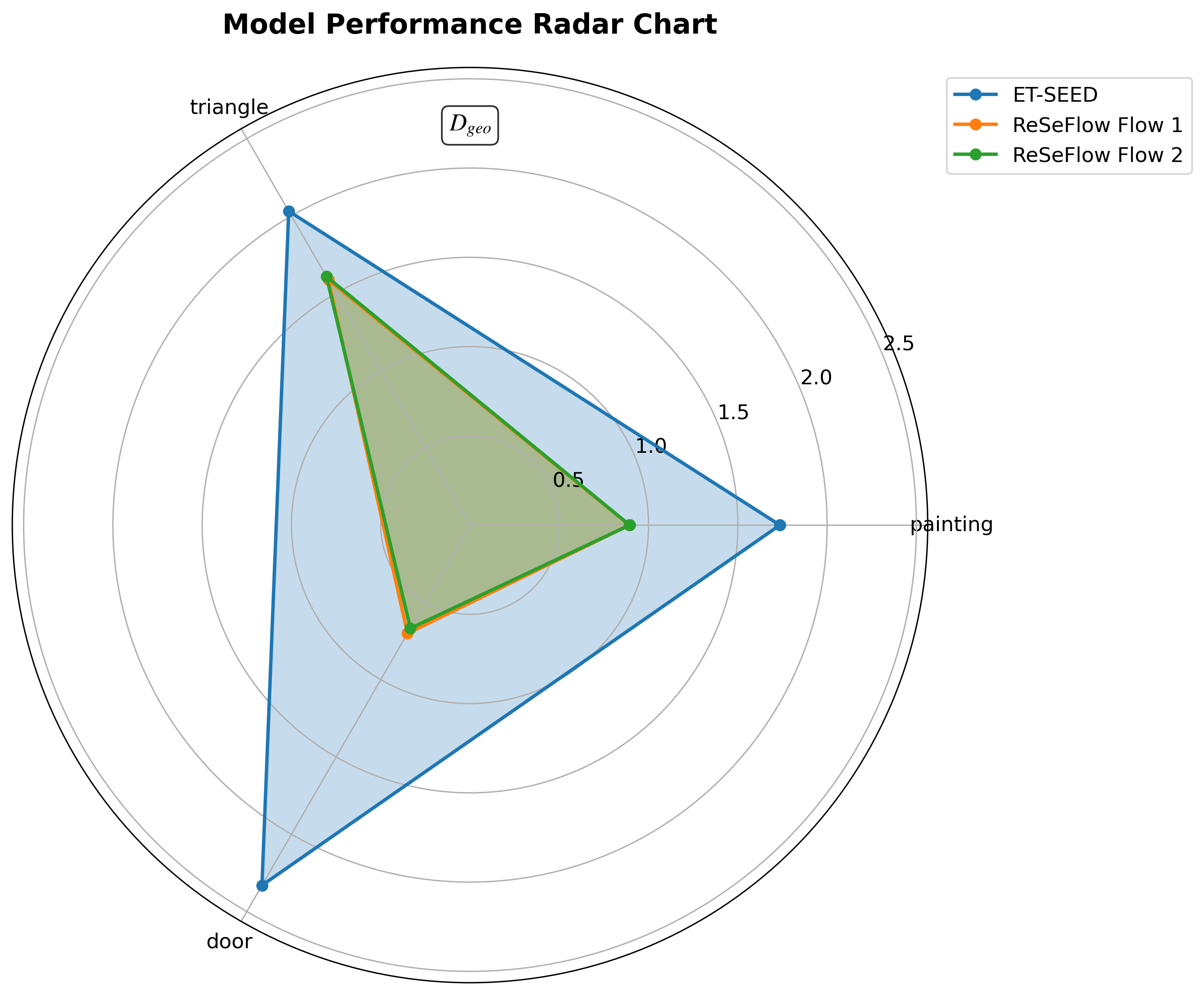}
    \caption{Model Performance Radar Chart. This figure demonstrates the $D_{geo}$ performance comparison of three models (ET-SEED, ReSeFlow Flow 1, and ReSeFlow Flow 2) across three manipulation tasks, i.e., rotating \textit{triangle}, \textit{door} opening, and \textit{painting}. Each axis of the radar chart represents a scene, with greater distance from the center indicating higher $D_{geo}$ values (worse performance). The results for ReSeFlow are based on the average of 10 runs with inference step set as 1.}
    \label{fig:model_radar_chart}
\end{figure}

Recently, the attention of many researchers are attracted by the \ac{se3}-equivariant diffusion models for robotic manipulation action generation in both target-pose manner \cite{ryu2024diffusion} and trajectory-planning manner \cite{hsu2025spot,lv2025spatial} thanks to the learning efficiency with limited data availability. The foregoing studies indicate three open questions, which motivates this work. First, capturing cross-modal trajectory-level robotic action sequences with a long horizon in \ac{se3} demands models to operate on the complex geometry of 3D space. Second, achieving a data-efficient learning process with only a few demonstrations but generalizing on diverse viewpoints, object poses, and scenario configurations requires the model to encode geometric perception and remain sufficient enough to represent potentially varying strategies. Third, a least-computational inference mechanism is essential for applying the policy generation process to the real-world case.

To answer the questions listed above, we are inspired by latest explorations in fast and robust generative models via rectified flow \cite{liu2022flow,yan2024perflow} with the application to object pose estimation and robotic policy inference \cite{sun2025rectified,ma20253d}. In this paper, we propose a two-branch framework \ac{reseflow} that inserts the rectification to the \ac{se3}-equivariant robotic trajectory-based policy learning flows. A \ac{rf} is leveraged in \ac{se3} to provide geometry-preserving and fast pose inference. Concretely, our contributions are summarized as follows,
\begin{enumerate}
    \item the design of a unified generative architecture that combines \ac{rf} with \ac{se3}-equivariant diffusion to generate action sequences for long-horizon robotic manipulation with the vision input conditions,
    \item the \ac{se3}-aware configuration that enforces geometric equivariance in the perceptional observation and enables a higher robustness and generalization across rigid-body motions and unseen scenarios,
    \item a data-efficient learning pipeline that allows the training of the model with only a few demonstrations,
    \item an inference-efficient generation process enabled by a multi-stage training strategy, i.e., diffusion-manner model training for the \ac{reseflow} Flow 1, followed by a reflow training for Flow 2 by using the generated data from Flow 1, and
    \item empirical validations on challenging simulated manipulation tasks, i.e., \textit{rotating triangle}, \textit{door opening}, and \textit{painting}, demonstrating improved generalization to robotic policy generation with significantly less inference steps compared with the baseline models, as shown in Fig. \ref{fig:model_radar_chart}.
\end{enumerate}

The rest of the paper is organized as follows. Section \ref{sec:related} reviews backgrounds on diffsuion-based policy generation, \ac{se3}-equivariance for action learning, and rectified flow. Section \ref{sec:pre} outlines mathematic bases for our proposed method and the conceptualization of the baseline methods. Section \ref{sec:methods} details the proposed \ac{reseflow} framework and its training objectives. Section \ref{sec:exp} presents experimental setup, results, discussion and limitations. The conclusion is drawn in Section \ref{sec:conc} with potential extensions.


\section{Related Work}
\label{sec:related}
\subsection{Diffusion Policies}

The diffusion-based robotic action generation models are boosted by the proposal of \ac{dp} \cite{chi2023diffusion} linking the image-based observation to action policy via a diffusion, with its followers, e.g., the 3D Diffusion Policy (DP3) \cite{ze20243d} utilizing the point cloud information input and Hierarchical Diffusion Policy
(HDP) \cite{ma2024hierarchical} targeted at multi-task manipulation with high-level agent. The policy generated by diffusion is also taking place in multi-modal robotic manipulation, i.e., the Vision-Language-Action (VLA) models. These models are aimed at understanding the image and text simultaneously and generating the controllable behaviors of robots \cite{sapkota2025vision}. TinyVLA \cite{wen2025tinyvla} provides a data-efficient policy generation with a fast inference. Although the aforementioned methods show potential performance in the manipulation task, the data demand and generality of models are still the main drawbacks of Euclidean diffusion.


\subsection{Equivariant Action Learning}
The \ac{se3} stands out in the robotic family as it involves all Euclidean motions and makes manipulation simple and solvable \cite{brockett2005robotic}. Among the sub-manifolds of \ac{se3}, the \textit{Lie algebra} $\mathfrak{se}(3)$ of \ac{se3} brings the Lie Group a success in the manipulation task \cite{wu2016inversion}. Currently, researchers are achieving the equivariant action generator by encoding the advantageous symmetric representation into the diffusion models. Diffusion-\ac{edf} are proposed on the \ac{se3} manifold to generate the target action in a non-trajectory manner by using a bi-equivariant score matching model \cite{ryu2024diffusion}. To make the generation into trajectory-level, \cite{yang2024equibot} and \cite{tie2024seed} propose the diffusion-based policy models in \ac{se3}. These models hold their competitive advantages with data efficiency as only a few demonstrations are required in the training phase for obtaining a comparable model. The inference efficiency is still the main challenge as a result of the iterative manner of the diffusion-based models.


\subsection{Rectified Flow}
Targeted at discovering the straightforward paths of \ac{ode} models, the rectification is proposed to make the generation flow in a higher efficiency even with a single inference step \cite{liu2022flow}, i.e., \ac{rf}. Recently, the principles are used in image generation \cite{zhu2024slimflow} and quality enhancement \cite{zhu2024flowie}, lidar data generation \cite{nakashima2024fast}, etc. Only a few advances are aimed at exploring the performance on the policy generation by using \ac{rf}. In order to achieve the efficient robot learning process, \cite{xuerecflow} solves the drawbacks of multi-step denoising process of diffusion for policy generation by using a \ac{rf}. However, the research in \ac{se3}-equivariant \ac{rf} for robotic policy learning still remains open. This paper moves forward with ET-SEED and fulfills the aforementioned gap by proposing a both data and inference-efficient \ac{reseflow}. 

\begin{figure*}[!t]
  \centering
  \begin{subfigure}{0.33\textwidth}
    \centering
    \includegraphics[height=1.8in]{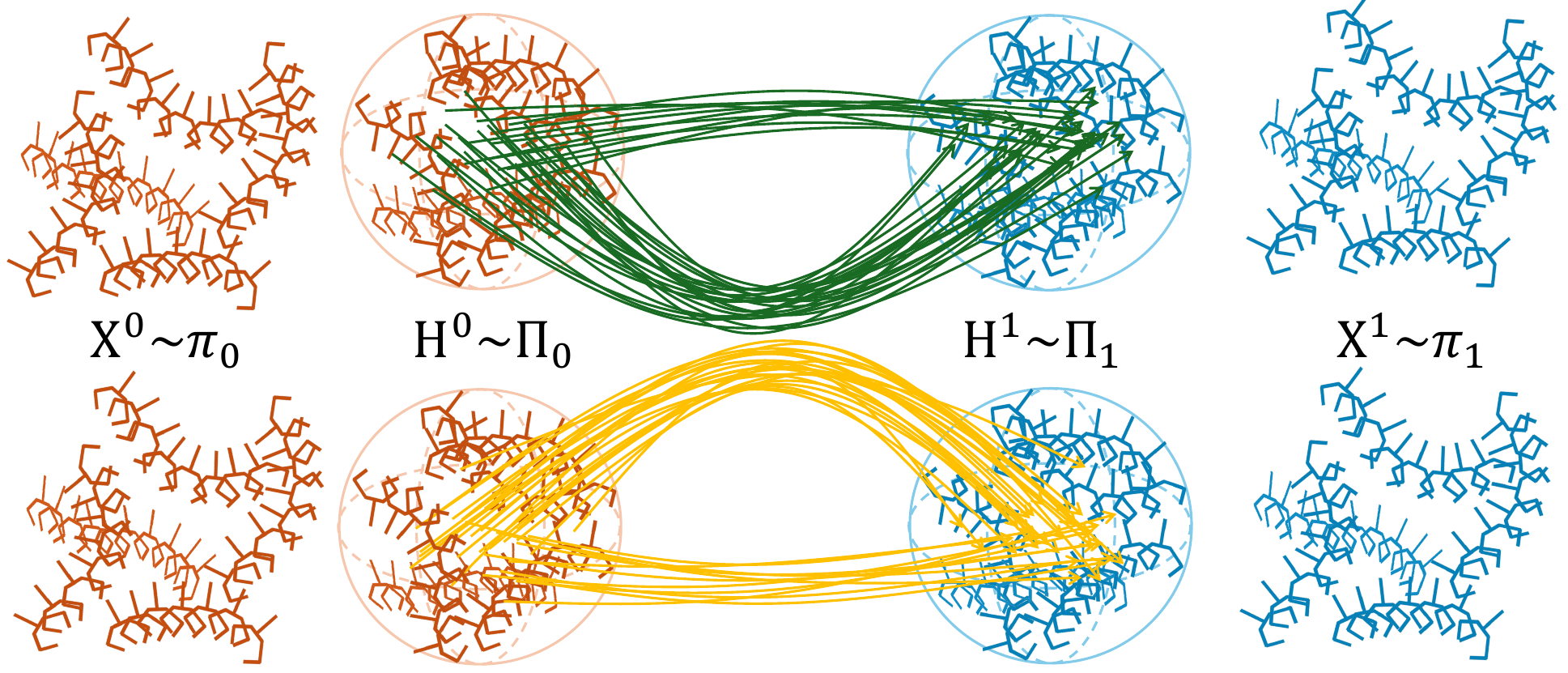}
    \caption{\ac{reseflow} Flow 1}
    \label{fig:flow1}
  \end{subfigure}\hfill
  \begin{subfigure}{0.33\textwidth}
    \centering
    \includegraphics[height=1.8in]{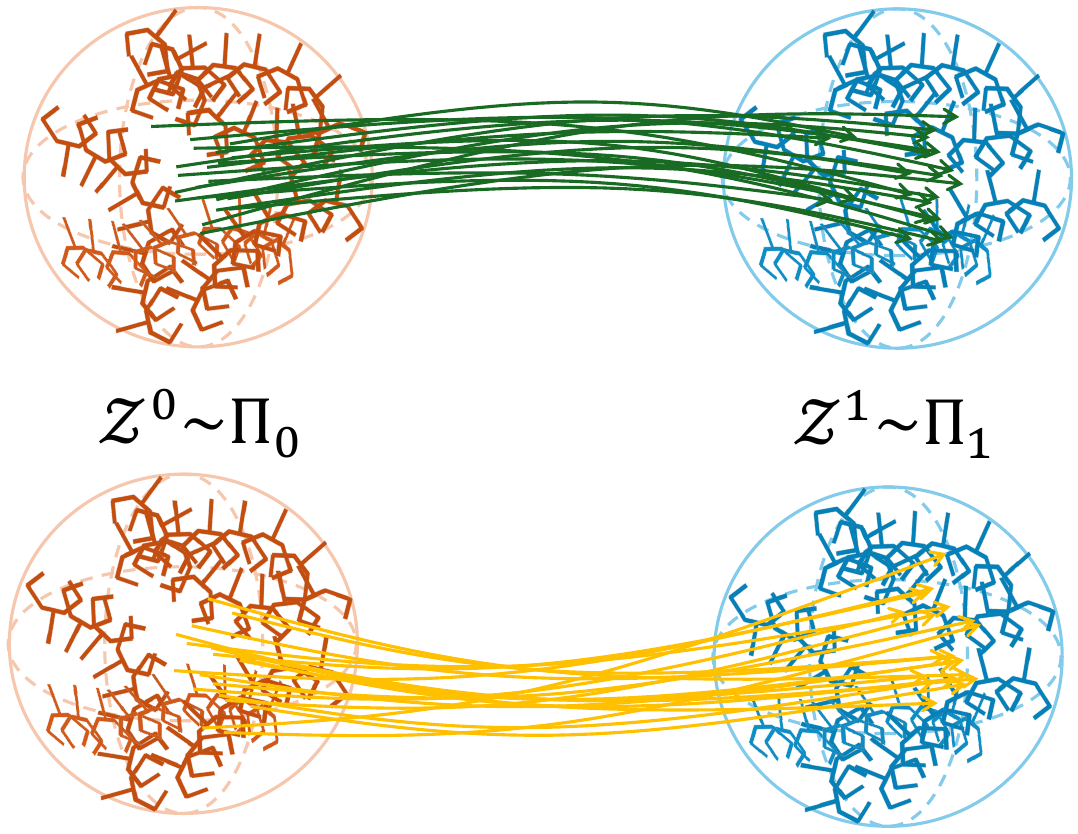}
    \caption{\ac{reseflow} Flow 2}
    \label{fig:flow2}
  \end{subfigure}
  \caption{The diagram of visualizing the \ac{reseflow} Flow 1 and Flow 2 in \ac{se3}. In Flow 1, the rectified flow $\mathcal{Z}$ is induced by the training data $\mathbf{H}^0$ and $\mathbf{H}^1$ in sequential manner, and in Flow 2, the reflow $\mathcal{Z}^\prime$ is induced by both the training data and the generated data from $\mathcal{Z}^0$ and $\mathcal{Z}^1$ after Flow 1.}
  \label{fig:overall}
\end{figure*}

\section{Preliminaries}
\label{sec:pre}
\subsection{Mathematic Base and \ac{se3}}
As a Lie group, \ac{se3} denotes the collection of 3D rigid-body transformations involving rotations and translations as

\begin{equation}
    SE(3) 
= \Bigl\{\,
(\mathbf{R}, \mathbf{t})
\,\Bigm\vert\,
\mathbf{R} \in SO(3), 
\,\mathbf{t} \in \mathbb{R}^3
\Bigr\},
\label{equ:se3}
\end{equation}
 where $\mathbf{R}$ is a $3 \times 3$ rotation matrix and $\mathbf{t}$ is a translation vector in 3D. Each element of \ac{se3} indicates a transformation, and the \textit{Lie algebra} $\mathfrak{se}(3)$ is a vector space. The \textit{expeotential map}, denoted as $\mathsf{exp}$, is used to map the \textit{Lie algebra} to \ac{se3}, i.e., $\mathsf{exp}: \mathfrak{se}(3) \longrightarrow SE(3)$ locally and diffeomorphically \cite{wu2016inversion}.

\subsection{Baseline Methods: ET-SEED}
\label{sec:pre_etseed}
\ac{etseed} generates a sequence of actions in \ac{se3}, $A^K = \left[\mathbf{H}^K_1, \mathbf{H}^K_2, ..., \mathbf{H}^K_{L}\right]$ via an equivariant diffusion model with the diffusion process from the noise-free actions $\mathbf{H}^0_i \in SE(3)$ into noisy actions $\mathbf{H}^K_i$ and denoising process vice versa. In \cite{tie2024seed}, the prediction length of the actions, i.e., the horizon, is denoted as $L$, and the diffusion steps are defined as $K$. An interpolation-based \ac{se3} diffusion model is used and the action at noise step $k$ is represented as 
\begin{equation}
    \mathbf{H}^k = \exp{\left( \gamma \sqrt{1-\bar{\alpha}_t} \varepsilon \right)} \mathcal{F}\left( \sqrt{\bar{\alpha}_t}; \mathbf{H}^0, \mathbb{H} \right), \quad \varepsilon \sim \mathcal{N}(0, \boldsymbol{\text{I}}),
\end{equation}
where $\mathbb{H}$ is the identify transformation in \ac{se3}, and $\varepsilon$ is the noise generated by the \ac{se3}-equivariant model.

\subsection{Rectified Flow}
\label{sec:pre_rf}
The generative flow is rectified by forcing the flow direction via a linear path diven by a force $v$. Given two empirical observations of $X_0 \sim \pi_0$ and $X_1 \sim \pi_1$, the induced \ac{rf} $Z_t$ with the force driving direction from $X_0$ to $X_1$, i.e., $(X_1-X_0)$, is defined by an \ac{ode} on time $t \in [0, 1)$ \cite{liu2022flow} as
\begin{equation}
    dZ_t = v(Z_t,t)dt.
\end{equation}
With a linear interpolation $X_t = tX_1 + (1-t) X_0$, \ac{se3} achieves the linear path $X_0 \longrightarrow X_1$ by solving
\begin{equation}
    \min_v \int_0^1 \mathbb{E} \left[\left\Vert(X_1 - X_0) - v(X_t, t)\right\Vert^2\right] dt.
\end{equation}
As \cite{liu2022flow} indicates the rectification procedure enables more straightforward path of generative process, the application of it to \ac{etseed} has the potential to reduce the inference costs.

\section{Method}
\label{sec:methods}
Aiming to achieve both data- and inference-efficient trajectory-level policy generation for robotic manipulation, we propose the \ac{reseflow}, a unified framework that rectifies \ac{se3}-equivariant policy learning flows. Fig. \ref{fig:overall} visualizes the learning flows for the sequential actions including Flow 1 and Flow 2 in \ac{se3}. 


\subsection{Definition of \ac{ode} for \ac{rf} in \ac{se3}}
An action trajectory with a prediction horizon $L$ is defined as a rectified sequence of poses in \ac{se3} 
\begin{equation}
    \mathcal{Z} = (\boldsymbol{\mathcal{Z}}_1, \boldsymbol{\mathcal{Z}}_2, \dots, \boldsymbol{\mathcal{Z}}_{L}), \quad \boldsymbol{\mathcal{Z}}_i \in SE(3).
\end{equation}
In order to explicitize the \ac{rf} in \ac{se3}, we define two distributions $\Pi_0, \Pi_1$ on the $\mathfrak{se}(3)$ of \ac{se3}. We assume $\mathcal{Z}_i^0 \sim \Pi_0$ and $\mathcal{Z}_i^1 \sim \Pi_1$. Equivalently, these distributions can be considered on \(\mathfrak{se}(3)\) via the pushforward of the \(\log\) map. The \ac{ode} in \ac{se3} is defined on the \textit{Lie algebra} $\mathfrak{se}(3)$. Let $\hat{\xi}_\theta(\mathcal{Z}; \mathcal{Z}^0, \mathcal{Z}^1) \in \mathfrak{se}(3)$ be the $4 \times 4$ twist matrix corresponding to a six-dimensional twist \((\omega, \rho)\) (angular and translational components). Denote the force-driving function of the induced flow in \ac{se3} as
\begin{equation}
   \hat{\xi}_\theta(\mathcal{Z}; \mathcal{Z}^0, \mathcal{Z}^1) =
\begin{bmatrix}
[\omega]_\times & \rho \\
0 & 0
\end{bmatrix}, 
\end{equation}
where \([\omega]_\times \in \mathbb{R}^{3\times 3}\) is the skew-symmetric matrix in \(\mathfrak{so}(3)\).

The \ac{ode} of \ac{rf} in \ac{se3} is
\begin{equation}
\frac{d}{dt} \mathcal{Z}(t) = \mathcal{Z}(t) \, \hat{\xi}_\theta\big(\mathcal{Z}(t); \mathcal{Z}^0, \mathcal{Z}^1\big),
\label{eq:rf-ode}
\end{equation}
with the initial condition at $t=0$, i.e., $\mathcal{Z}(0) = \mathcal{Z}^0$, and the flow end at $t=1$, i.e., $\mathcal{Z}(1) = \mathcal{Z}^1$. The drift function $\hat{\xi}_\theta$ is designed to transport $\mathcal{Z}^0$ to $\mathcal{Z}^1$ with a rectified flow in \ac{se3}.

We adopt standard Lie-theoretic maps on \(SE(3)\). Let \(\log: SE(3) \to \mathfrak{se}(3)\) and \(\mathrm{Exp}: \mathfrak{se}(3) \to SE(3)\) denote the matrix logarithm and exponential. For a twist \(\xi = (\omega, \rho) \in \mathbb{R}^6\), the hat operator maps to \(\hat{\xi} = \begin{bmatrix} [\omega]_\times & \rho \\ 0 & 0 \end{bmatrix} \in \mathfrak{se}(3)\), and the vee operator \((\cdot)^{\vee}\) is its inverse from \(\mathfrak{se}(3)\) to \(\mathbb{R}^6\). For two poses \(X, Y \in SE(3)\), we define their geodesic difference as
\begin{equation}
X \ominus Y \;\coloneqq\; \bigl(\log(X\,Y^{-1})\bigr)^{\vee} \in \mathbb{R}^6.
\end{equation}
Accordingly, we denote the straight-path displacement in the Lie algebra by \(\Delta^{\vee} = ( \mathbf{H}^1 \ominus \mathbf{H}^0 )\).
The geodesic interpolation in \(SE(3)\) is
\begin{equation}
\mathbf{H}^t \;=\; \mathrm{Exp}\!\Bigl(t\,\log(\mathbf{H}^1 (\mathbf{H}^0)^{-1})\Bigr)\,\mathbf{H}^0.
\end{equation}
which yields a straight line in \(\mathfrak{se}(3)\) under the \(\log\) map and is consistent with the rectified-flow formulation.

\subsection{\ac{se3}-equivariant Flow 1 and Flow 2}
Given the observation sequences $\mathbf{H}^0 \sim \Pi_0$ and $\mathbf{H}^1 \sim \Pi_1$ in \ac{se3}, following the vanilla \ac{rf}, we aim to disclose the least-computational (geodesic) path of $\Pi_0 \longrightarrow \Pi_1$ by solving
\begin{equation}
\min_{\hat{\xi}_\theta} \int_0^1
\mathbb{E}\!\left[
\left\lVert (\mathbf{H}^1 \ominus \mathbf{H}^0) - \hat{\xi}_\theta(\mathbf{H}^t, t) \right\rVert_2^2
\right] \, dt.
\end{equation}
\begin{equation}
\mathbf{H}^t \;=\;
\mathrm{Exp}\!\big(t\,\log(\mathbf{H}^1 (\mathbf{H}^0)^{-1})\big)\,\mathbf{H}^0.
\end{equation}
where $\mathbf{H}^t$ denotes the geodesic interpolation in $SE(3)$ given above, inducing the rectified Flow~1 as $\mathcal{Z}$. The re-application of rectification to the rectified flow can significantly accelerate straightening the flow path \cite{kim2024simple}. Therefore, a reflow $\mathcal{Z}^\prime$ is induced iteratively by $\mathcal{Z}^0$ and $\mathcal{Z}^1$ with $\mathcal{Z}^t = \mathrm{Exp}\!\big(t\,\log(\mathcal{Z}^1 (\mathcal{Z}^0)^{-1})\big)\,\mathcal{Z}^0$, which is trained based on the generated data from Flow 1 and original training data. Flow 2 augments Flow 1 by disclosing a more straightforward alignment from Flow 1 prior. This two-flow pipeline enables a strong, geometry-aware and efficient policy generation that is robust to viewpoint and pose variations.

In practice, we minimize a Monte Carlo approximation of the continuous objective:
\begin{equation}
\begin{gathered}
\mathcal{L}_{\mathrm{F1}}(\theta) =
\mathbb{E}_{(\mathbf{H}^0,\mathbf{H}^1)\sim \mathcal{D},\, t\sim \mathcal{U}(0,1)}
\left[
\lVert (\mathbf{H}^1 \ominus \mathbf{H}^0) - \xi_\theta(\mathbf{H}^t, t) \rVert_2^2
\right].
\\[1.2ex]
\end{gathered}
\end{equation}

\vspace{0.5\baselineskip}

where \(\xi_\theta\) outputs a 6D twist and the norm is in \(\mathbb{R}^6\). The pose derivative follows the right-invariant ODE \( \frac{d}{dt}\mathcal{Z}(t) = \mathcal{Z}(t)\,\hat{\xi}_\theta(\mathcal{Z}(t); \mathcal{Z}^0, \mathcal{Z}^1) \). We discretize \(t\) by uniform sampling per minibatch.

For Flow~2 (reflow), we synthesize auxiliary training pairs using the Flow~1 checkpoint and mix them with original demonstrations at ratio \(\rho\). Concretely, we draw \(t \sim \mathcal{U}(0,1)\), construct straight-path samples in \(\mathfrak{se}(3)\) between data and a stochastic endpoint (e.g., isotropic Gaussian in \(\mathfrak{se}(3)\) or a forward pass of Flow~1), and supervise the same vector-field target:
\begin{equation}
\begin{aligned}
\mathcal{L}_{\mathrm{F2}}(\theta)
&=
\mathbb{E}
\Bigl[
\bigl\| (\mathbf{H}^1 \ominus \mathbf{H}^0) - \xi_\theta(\mathbf{H}^t, t)\bigr\|_2^2
\Bigr].
\end{aligned}
\end{equation}
\\
\noindent\textit{Minibatch mixing: } $\rho$:(reflow) + $(1-\rho)$:(original).
\\
We typically use a slightly smaller learning rate and a larger rectified-step budget in Flow~2 to refine the vector field learned in Flow~1.

\subsection{\ac{se3}-equivariant Drift Model}
As the proposed \ac{reseflow} operates in \ac{se3}, the drift model $\hat{\xi}_\theta$ is required to fulfill three goals including 1) generating robust and desired "force" limiting $\mathcal{Z}^0$ to $\mathcal{Z}^1$; 2) ensuring the operations of policy generation well-defined under rigid motions in \ac{se3} manifold; 3) crossing multi-domain knowledge, i.e., the visual perception to action space, and achieving a multi-modal functionality. In this work, we follow the backbone design used in \cite{tie2024seed} to achieve the desired drift model. 

Beyond the core design described above, this integrated \ac{reseflow} promotes data-efficient learning by incorporating the geometry-aware priors thanks to \ac{se3} manifold to reduce the reliance of the model on large-scale demonstrations, and enables fast policy inference with the application of rectification.

\subsection{Training and Inference Protocol}
We adopt a two-stage pipeline aligned with our quick-start implementation.
\\

\textbf{Flow 1 (original data)}

Sample \(t \sim \mathcal{U}(0,1)\), supervise on the straight path in \(\mathfrak{se}(3)\) using \(\Delta^{\vee}\), and train with a cosine learning-rate schedule and small batches for stability. We use a moderate rectified step budget and RK45 at test time.
\\

\textbf{Flow 2 (reflow mixture)}

Synthesize reflow pairs using the Flow~1 checkpoint and mix them with original demonstrations at ratio \(\rho\) (default \(\rho=0.5\)). Train with a slightly smaller learning rate and an increased rectified step budget to refine the vector field.
\\

\textbf{Inference and evaluation}

Integrate the learned \ac{ode} with high-accuracy solvers (e.g., RK45), and evaluate across different samplers and step budgets. We additionally measure multi-node performance and node-wise consistency.

\section{Experiment}
\label{sec:exp}
We validate the proposed \ac{reseflow} framework to emphasize geometry-aware generalization, multi-modality, and data efficiency under \ac{se3} symmetry as well as inference efficiency. We detail the tasks, simulation data generation, evaluation protocol, implementation, and results in this section.

\subsection{Experiment Setup}
\label{sec:exp_setup}

We undertake three typical table-top manipulation tasks, \textit{door opening}, \textit{painting}, and \textit{rotating triangle} in NVIDIA Isaac Sim platform that stresses long-horizon policy planning.

\begin{figure}[!t]
    \centering\
    \begin{tabular}{@{}cc@{}}
        \includegraphics[width=0.2\textwidth]{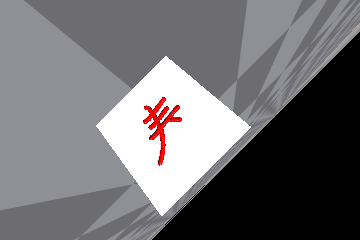} &
        \includegraphics[width=0.2\textwidth]{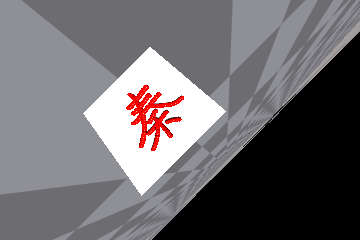} \\
        \multicolumn{2}{c}{\footnotesize\strut Painting} \\
        \includegraphics[width=0.2\textwidth]{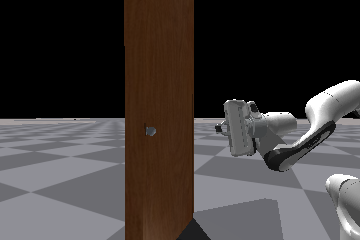} &
        \includegraphics[width=0.2\textwidth]{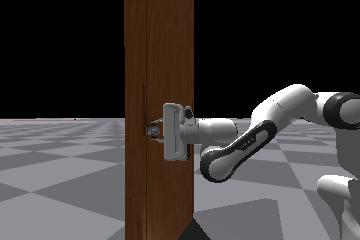} \\[2pt]
        \multicolumn{2}{c}{\footnotesize\strut Door opening} \\
        \includegraphics[width=0.2\textwidth]{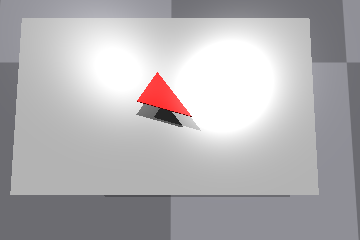} &
        \includegraphics[width=0.2\textwidth]{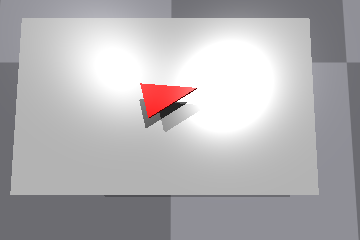} \\[2pt]
        \multicolumn{2}{c}{\footnotesize\strut Rotating Triangle}
    \end{tabular}
    \caption{Visualization of training and test data for three tasks: \textit{Painting}, \textit{Door opening}, and \textit{Rotating Triangle}.}
    \label{fig:tasks_vis}
\end{figure}

\textbf{Data Collection and Ground Truth} The trajectories are stored in structured numpy arrays containing point cloud observations, end-effector poses, and ground-truth actions. Specifically, the settings for each task are as follows,
\begin{enumerate}
    \item \textit{Painting}: 31 demonstrations with 256-point point clouds and 4$\times$4 action matrices, 
    \item \textit{Door opening}: 10 demonstrations with 128-point point clouds and 4$\times$4 action matrices, and 
    \item \textit{Rotating Triangle}: 500 demonstrations with 100-point point clouds and 4$\times$4 action matrices. 
\end{enumerate}
Each demonstration consists of 10 action steps with 4-dimensional action space (3D position + 1D gripper state) and 2-dimensional observation horizon. The simulation runs at 20Hz with physical timestep of 0.05s, ensuring realistic dynamics and contact interactions. Fig. \ref{fig:tasks_vis} shows the training and test data for three tasks: Painting, Door opening, and Rotating Triangle.


\textbf{Training Configuration} We employ a two-stage \ac{rf} training pipeline. In the first stage, we train a \ac{reseflow} Flow 1 by using the original demonstration data with the following hyperparameters: learning rate of 2$\times$10$^{-4}$, batch size of 1, 5000 epochs, and RK45 ODE solver for 100 integration steps. The second stage utilizes reflow data generation with 50\% synthetic data ratio, 3000 epochs, and learning rate of 8$\times$10$^{-5}$. The model architecture consists of a multi-node SE3ManiNet with 4 nodes, incorporating equivariant and invariant representations for robust manipulation learning.

\textbf{Hardware and Computational Details} All experiments are conducted on a single NVIDIA RTX 3090 GPU with 24GB VRAM. The evaluation is performed over 10 test runs per task with different random seeds (base seed: 3407), ensuring statistical significance of the results.

We aim to verify the performance of the proposed \ac{reseflow} by comparing the following methods,
\begin{itemize}
    \item \ac{etseed}: the vanilla equivariant policy generation method,
    \item \ac{reseflow} (Flow 1): the proposed method trained on the simulated data, and
    \item \ac{reseflow} (Flow 2): the proposed method with reflow trained on the generated data by Flow 1 and simulated data.
\end{itemize}

\subsection{Evaluation Protocol and Metrics}
We assess the generalization to the unseen object configurations and viewpoints with the pose error and trajectory quality, i.e., the translational and rotational errors of the action sequence.

\textbf{Geodesic Distance} We evaluate pose accuracy by using the geodesic distance in \ac{se3}. Let $T=(R,t)\in SE(3)$ be the predicted pose and $\hat{T}=(\hat{R},\hat{t})\in SE(3)$ the ground truth, with $R,\hat{R}\in SO(3)$ and $t,\hat{t}\in\mathbb{R}^3$.
The metric is
\begin{equation}
\label{eq:dgeo}
D_{\mathrm{geo}}(T,\hat{T})
= \sqrt{\left\|\,\mathrm{Log}\!\big(R^{\top}\hat{R}\big)\,\right\|_2^2
      + \left\|\,\hat{t}-t\,\right\|_2^2}\,,
\end{equation}
where $\mathrm{Log}:SO(3)\to\mathfrak{so}(3)$ is the matrix logarithm. Denoting
$\phi=\big(\mathrm{Log}(R^{\top}\hat{R})\big)^\vee\in\mathbb{R}^3$ as the axis–angle vector,
$\|\phi\|_2$ equals the rotation-angle error (radians), and $\|\hat{t}-t\|_2$ is the Euclidean translation error. The lower $D_{\mathrm{geo}}$, the better.

\textbf{Trajectory-level Visualization} For a trajectory $\{T_k\}_{k=1}^{L}$ with ground truth $\{\hat{T}_k\}_{k=1}^{L}$, we compute the per-action error
$D_{\mathrm{geo}}(T_k,\hat{T}_k)$ and (i) plot the per-action curve; (ii) report the trajectory-level average as
\begin{equation}
\label{eq:avg_dgeo}
\overline{D}_{\mathrm{geo}}
= \frac{1}{L}\sum_{k=1}^{L} D_{\mathrm{geo}}(T_k,\hat{T}_k),
\end{equation}
which is aggregated over 10 random seeds as mean $\pm1$ standard deviation.

\begin{figure*}[!t]
  \centering
  \begin{subfigure}{0.3\textwidth}
    \centering
    \includegraphics[width=\linewidth]{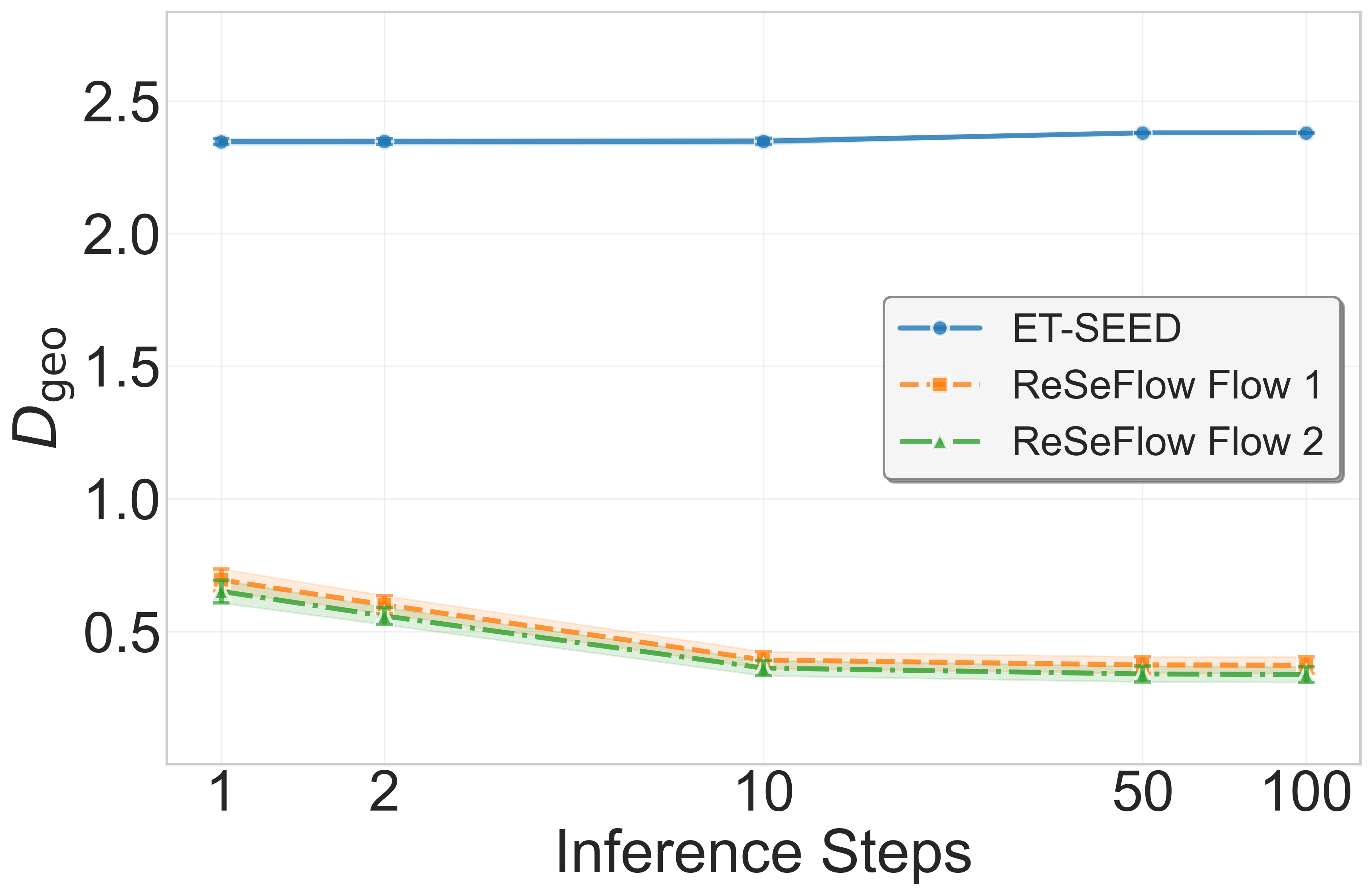}
    \caption{\textit{Door Opening}}
    \label{fig:infer-dooropen}
  \end{subfigure}\hfill
  \begin{subfigure}{0.3\textwidth}
    \centering
    \includegraphics[width=\linewidth]{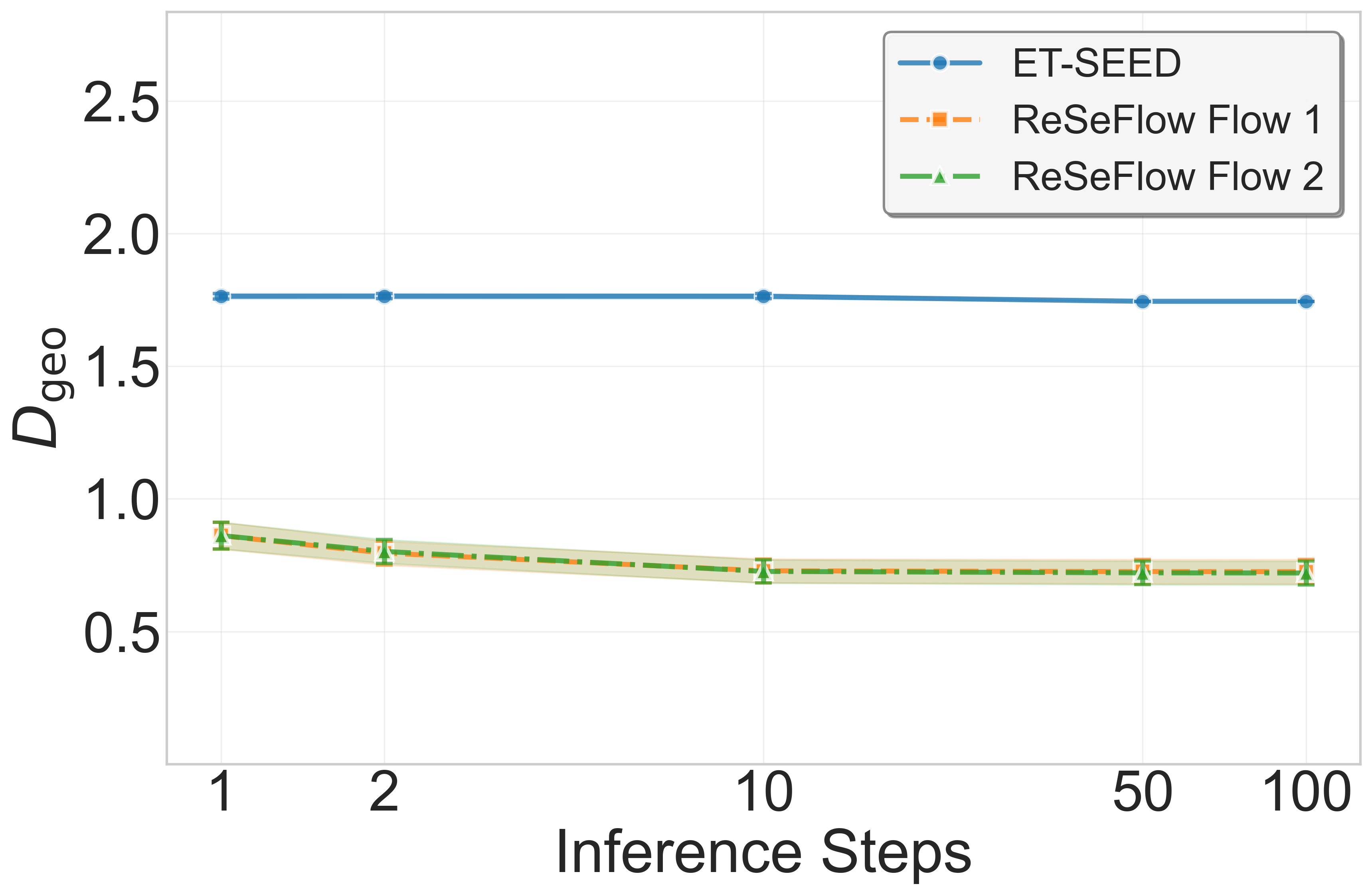}
    \caption{\textit{Painting}}
    \label{fig:infer-painting}
  \end{subfigure}\hfill
  \begin{subfigure}{0.3\textwidth}
    \centering
    \includegraphics[width=\linewidth]{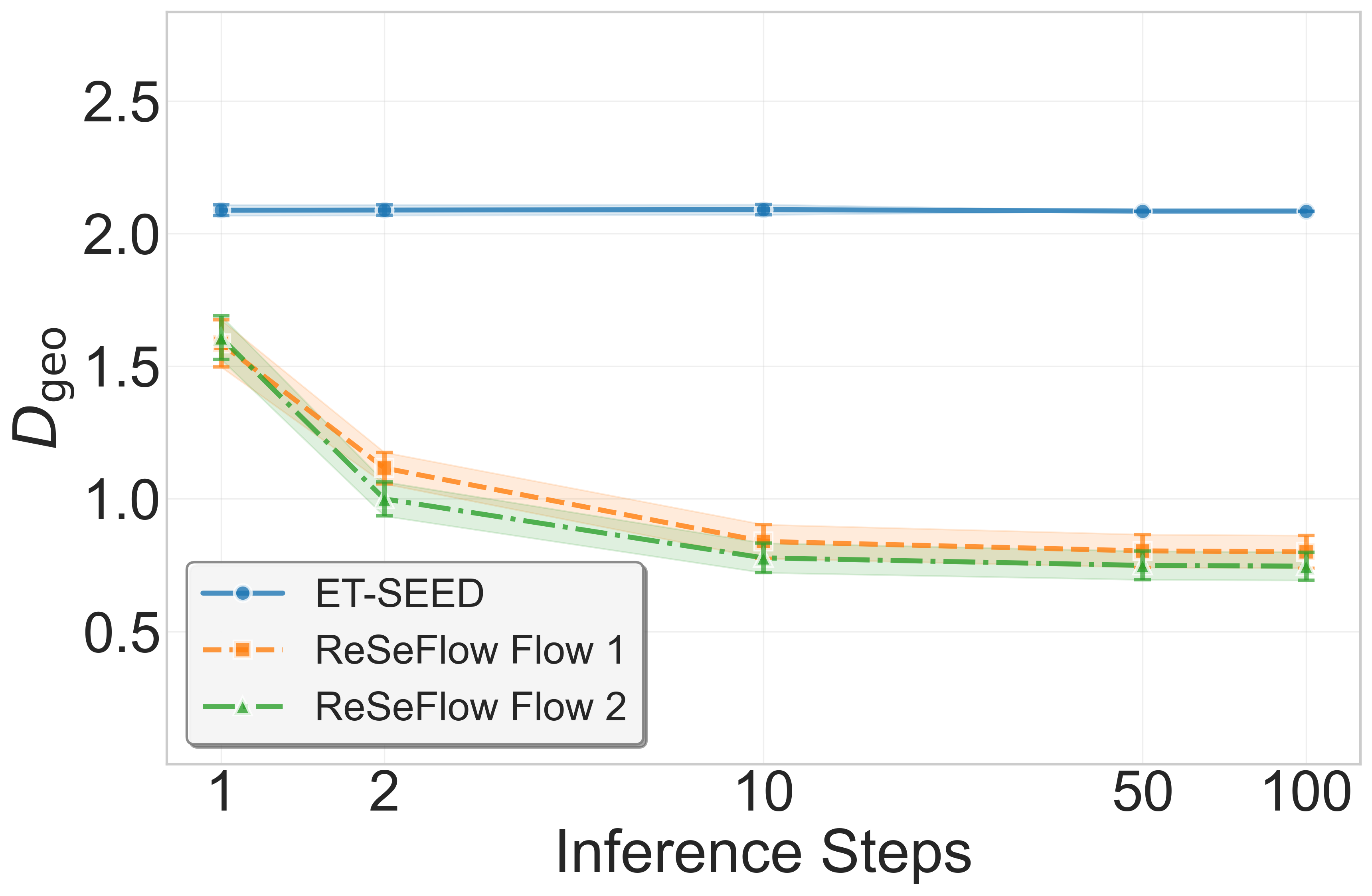}
    \caption{\textit{Rotating Triangle}}
    \label{fig:infer-triangle}
  \end{subfigure}
  \caption{Geodesic distance between predicted and ground-truth across inference steps 1, 2, 10, 50, and 100. From left to right: \textit{door opening}, \textit{painting}, and \textit{rotating triangle}.}
  \label{fig:geodesic-compare}
\end{figure*}

\begin{figure*}[!t]
  \centering
  \begin{subfigure}{0.25\textwidth}
    \centering
    \includegraphics[width=\linewidth]{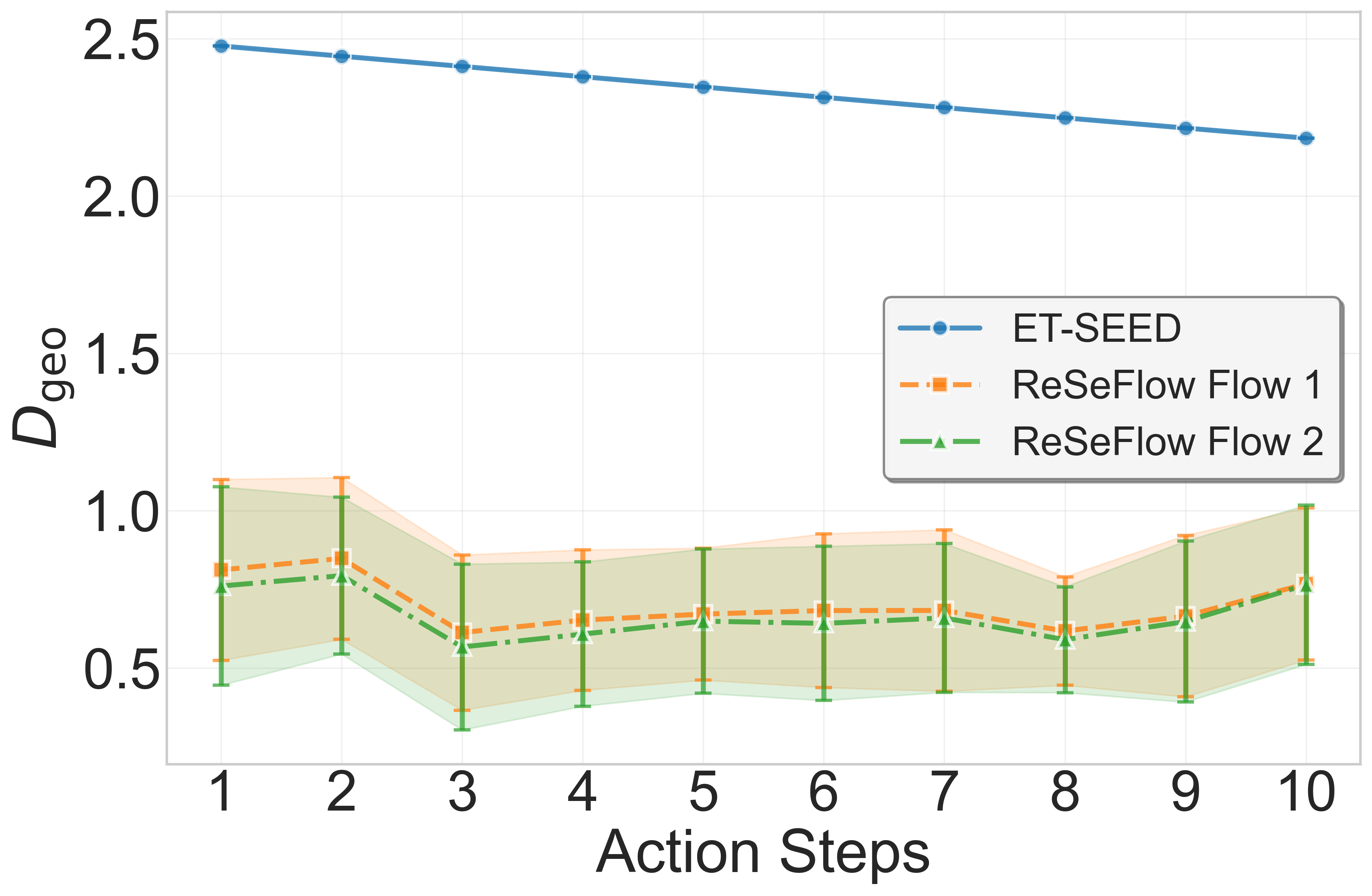}
    \caption{rf\_step = 1}
    \label{fig:door-rf1}
  \end{subfigure}\hfill
  \begin{subfigure}{0.25\textwidth}
    \centering
    \includegraphics[width=\linewidth]{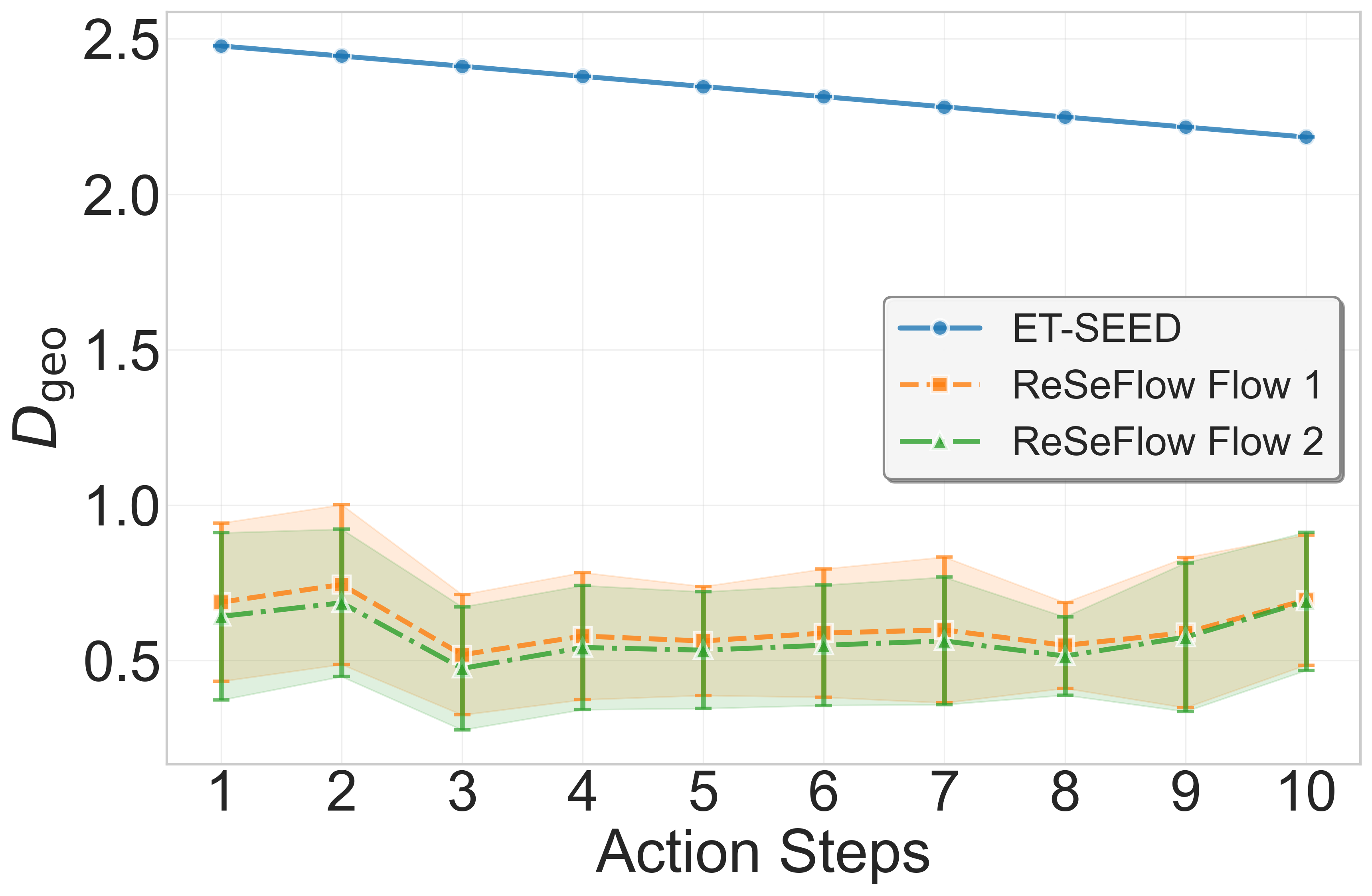}
    \caption{rf\_step = 2}
    \label{fig:door-rf2}
  \end{subfigure}\hfill
  \begin{subfigure}{0.25\textwidth}
    \centering
    \includegraphics[width=\linewidth]{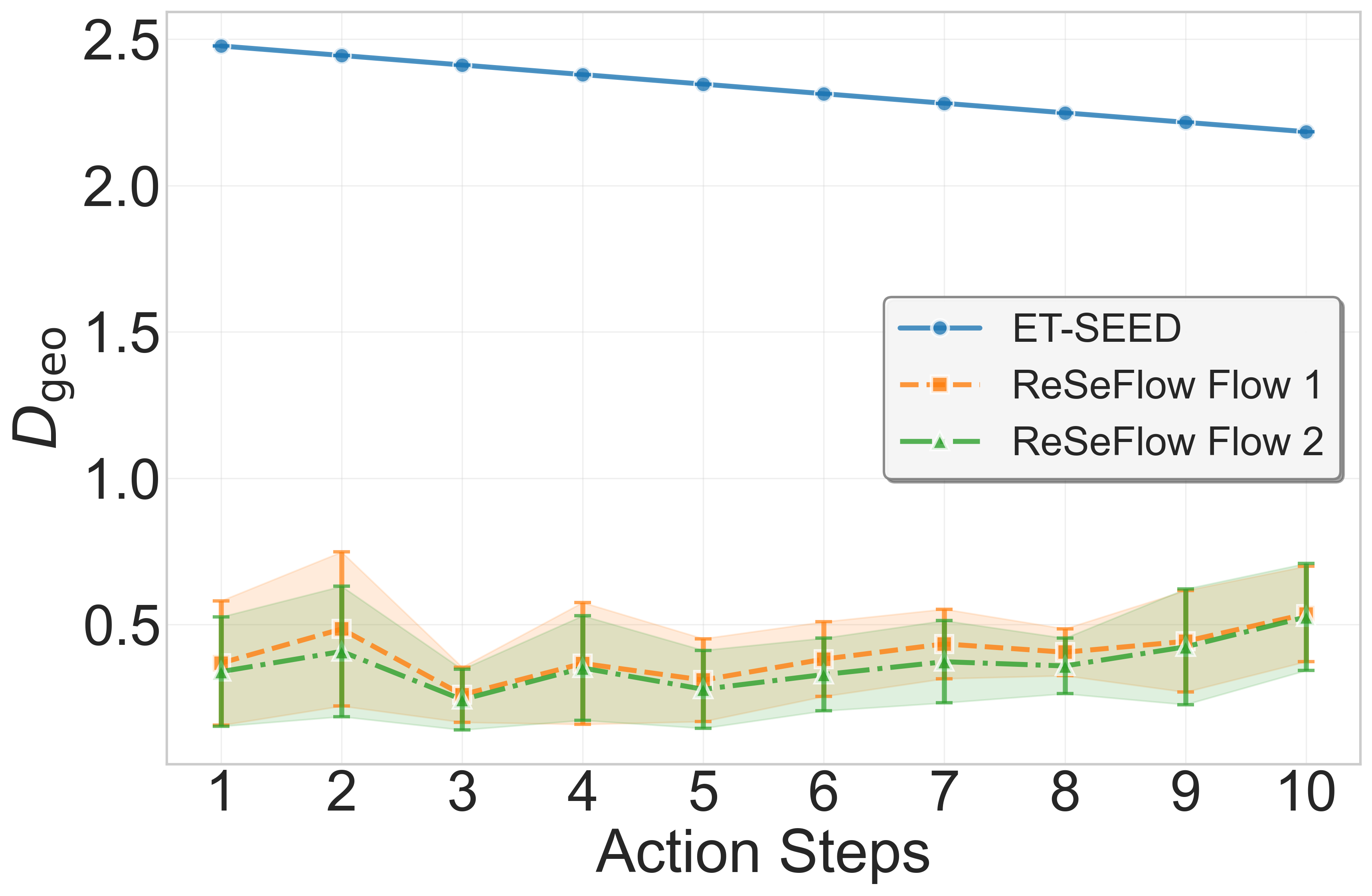}
    \caption{rf\_step = 50}
    \label{fig:door-rf50}
  \end{subfigure}\hfill
  \begin{subfigure}{0.25\textwidth}
    \centering
    \includegraphics[width=\linewidth]{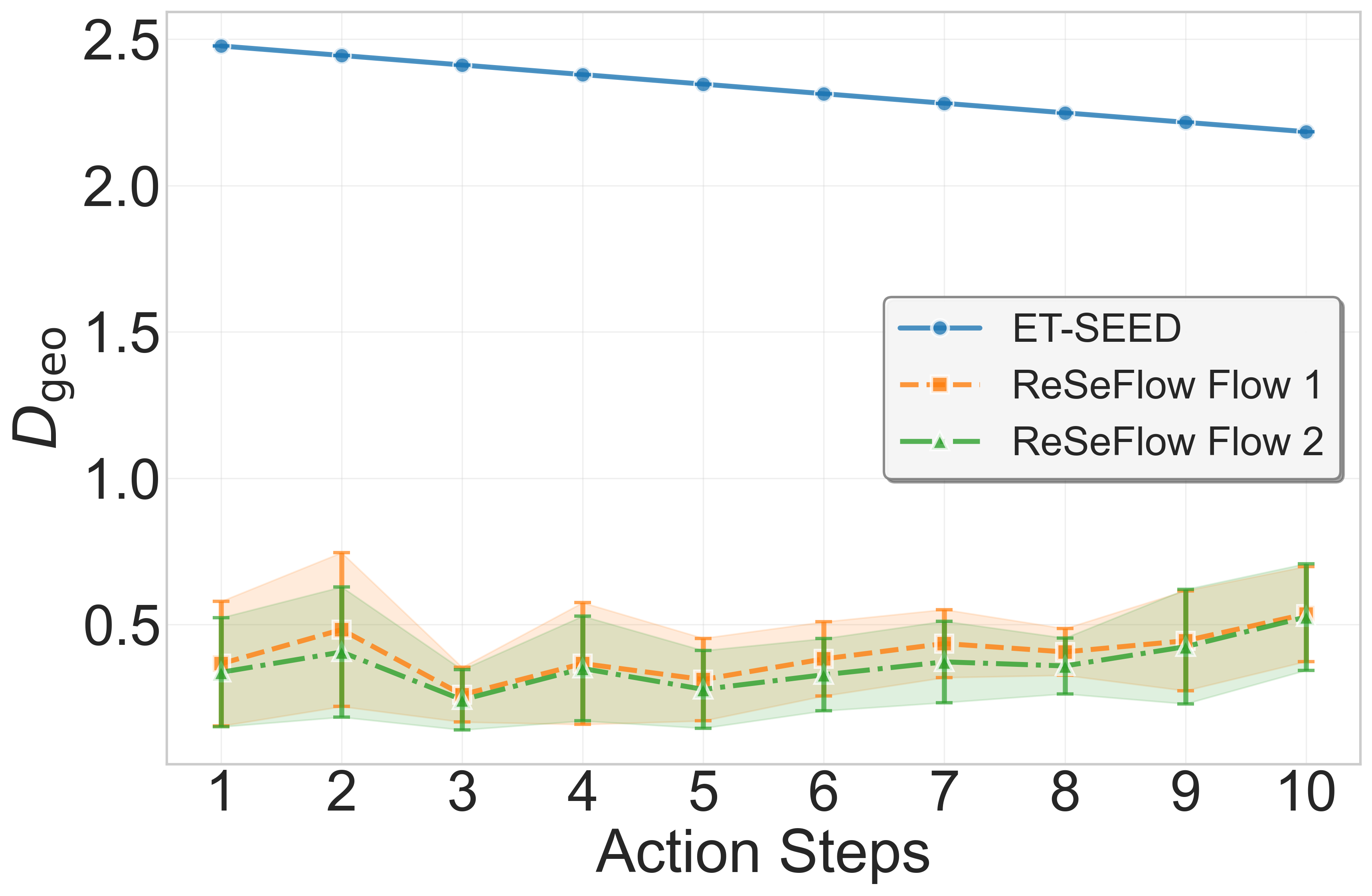}
    \caption{rf\_step = 100}
    \label{fig:door-rf100}
  \end{subfigure}

  \caption{\textit{Door opening} scene: distance over the full trajectory at \ac{reseflow} inference steps 1, 2, 50, and 100 (mean and variance over 10 seeds).}
  \label{fig:door-trajectory-distance}
\end{figure*}


\begin{figure*}[!t]
  \centering
  \begin{subfigure}{0.25\textwidth}
    \centering
    \includegraphics[width=\linewidth]{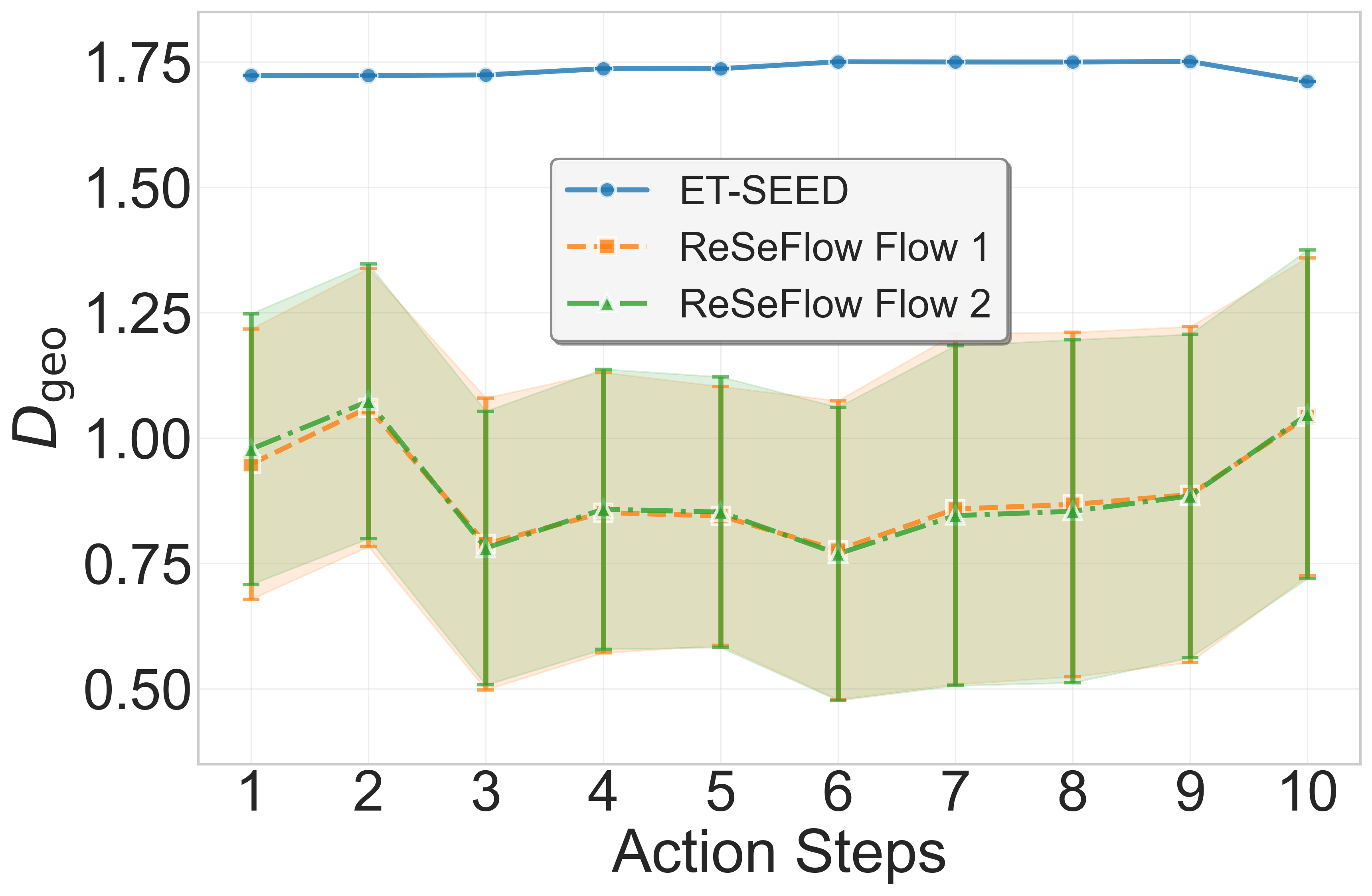}
    \caption{rf\_step = 1}
    \label{fig:paint-rf1}
  \end{subfigure}\hfill
  \begin{subfigure}{0.25\textwidth}
    \centering
    \includegraphics[width=\linewidth]{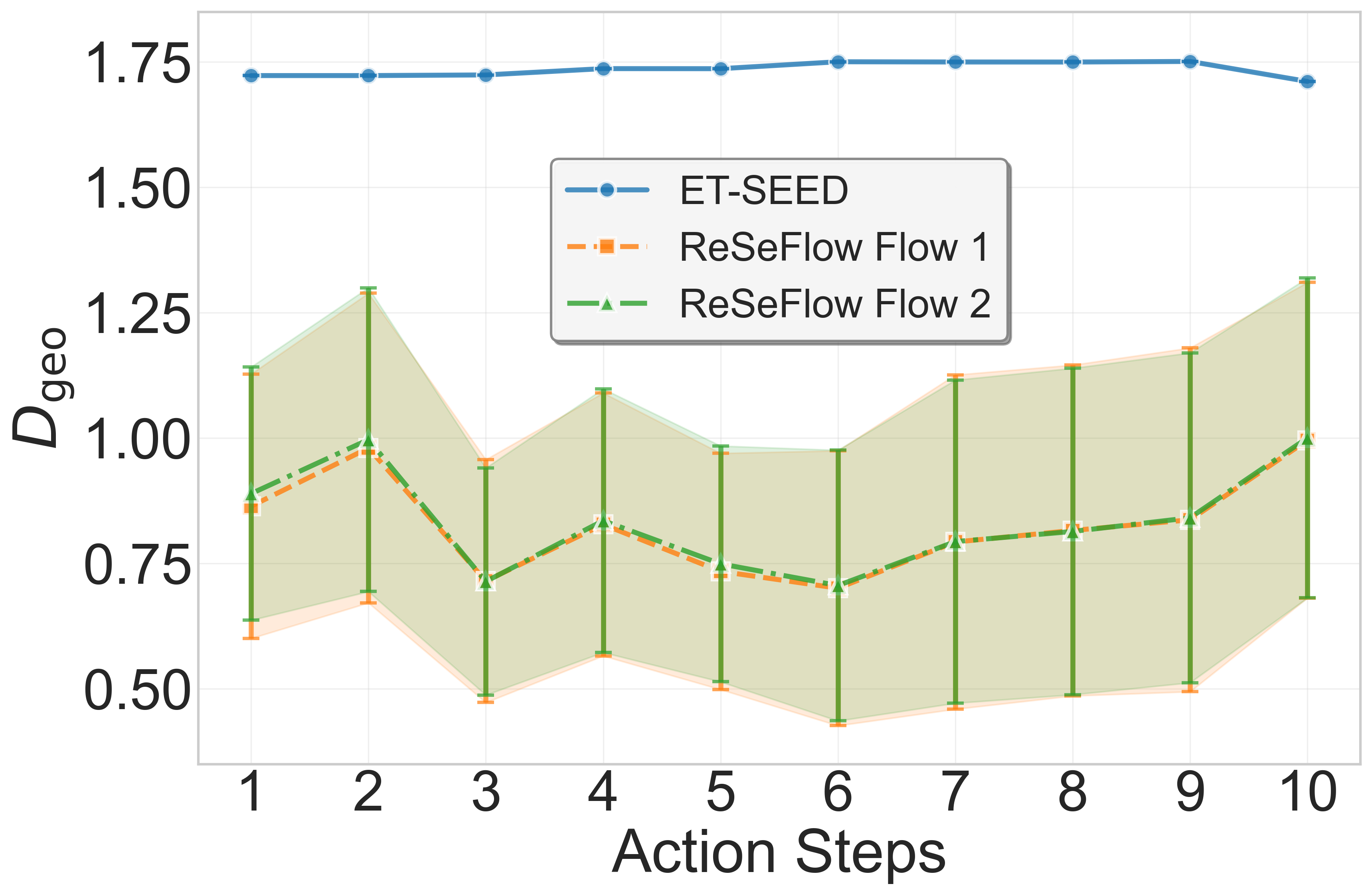}
    \caption{rf\_step = 2}
    \label{fig:paint-rf2}
  \end{subfigure}\hfill
  \begin{subfigure}{0.25\textwidth}
    \centering
    \includegraphics[width=\linewidth]{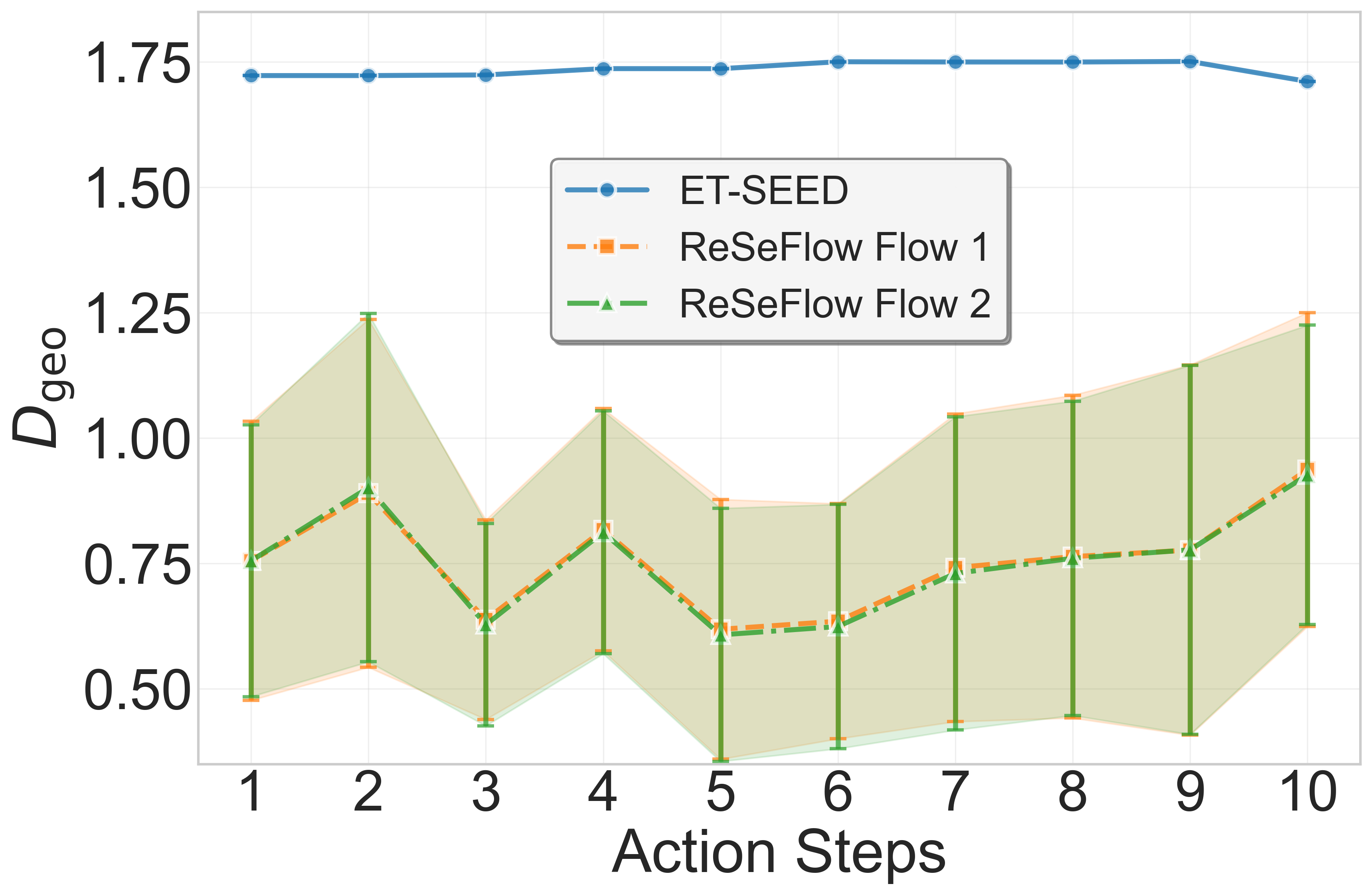}
    \caption{rf\_step = 50}
    \label{fig:paint-rf50}
  \end{subfigure}\hfill
  \begin{subfigure}{0.25\textwidth}
    \centering
    \includegraphics[width=\linewidth]{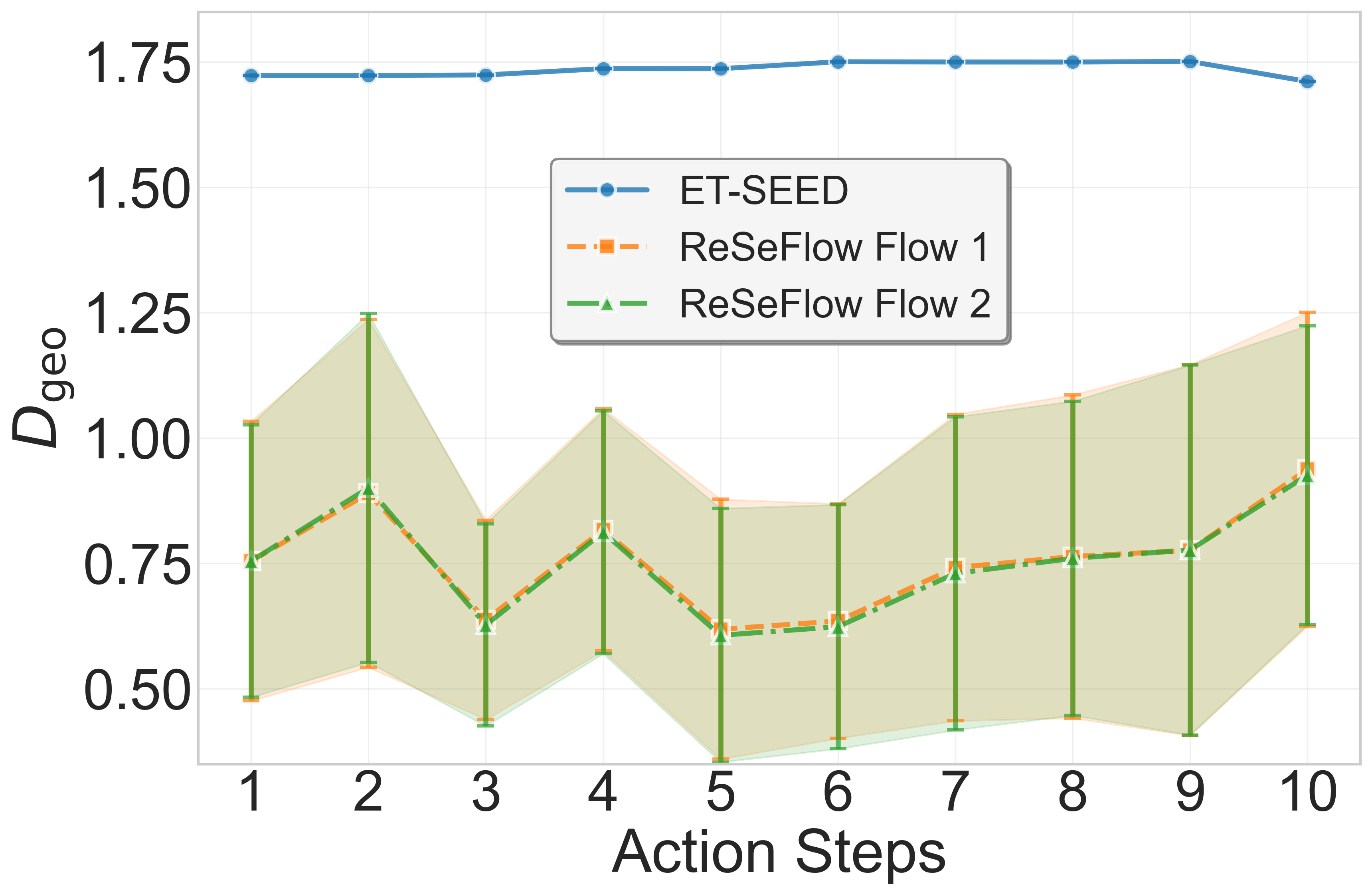}
    \caption{rf\_step = 100}
    \label{fig:paint-rf100}
  \end{subfigure}

  \caption{\textit{Painting} scene: distance over the full trajectory at ReSeFlow inference steps 1, 2, 50, and 100 (mean and variance over 10 seeds).}
  \label{fig:paint-trajectory-distance}
\end{figure*}

\begin{figure*}[!t]
  \centering
  \begin{subfigure}{0.25\textwidth}
    \centering
    \includegraphics[width=\linewidth]{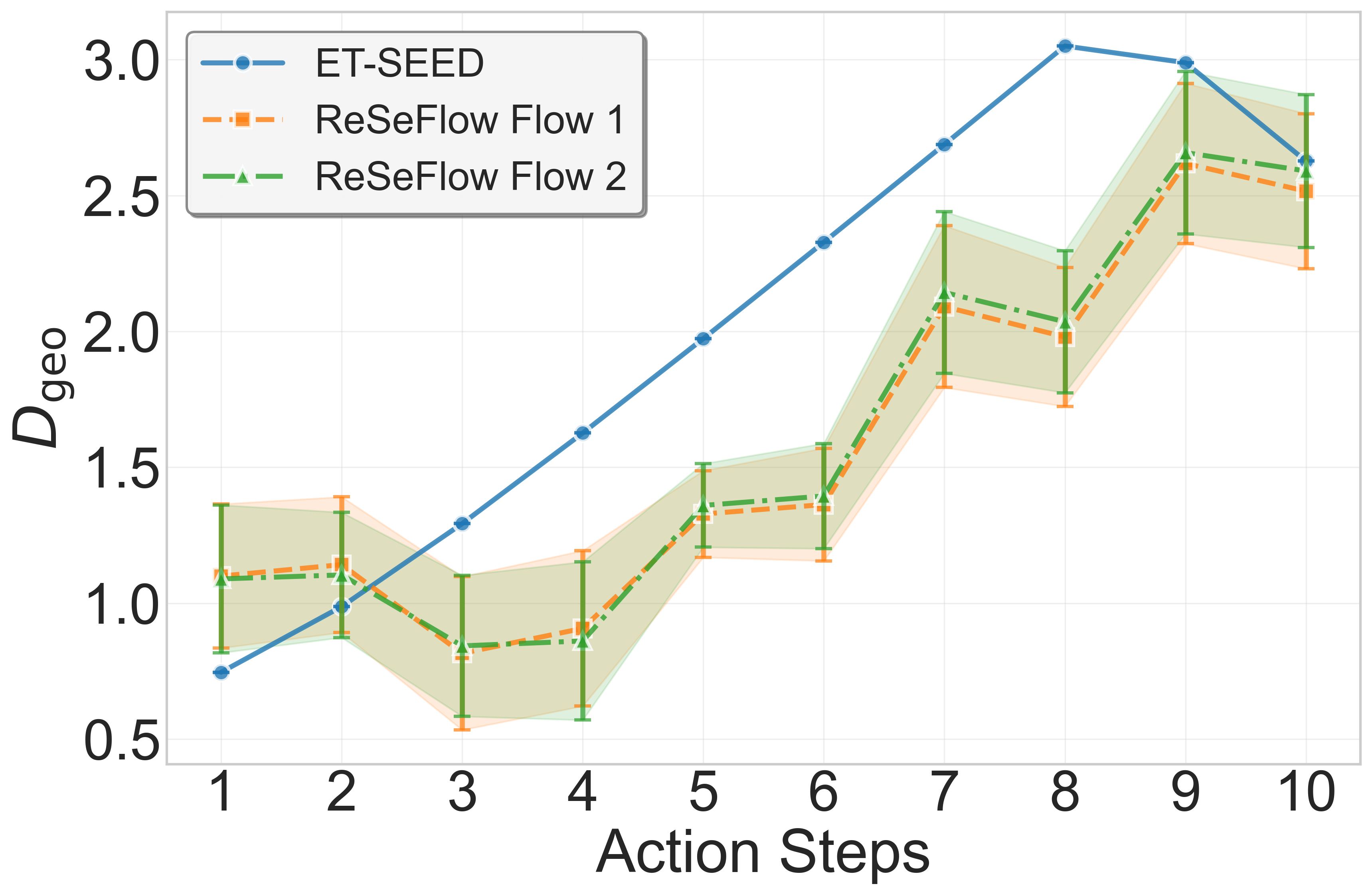}
    \caption{rf\_step = 1}
    \label{fig:tria-rf1}
  \end{subfigure}\hfill
  \begin{subfigure}{0.25\textwidth}
    \centering
    \includegraphics[width=\linewidth]{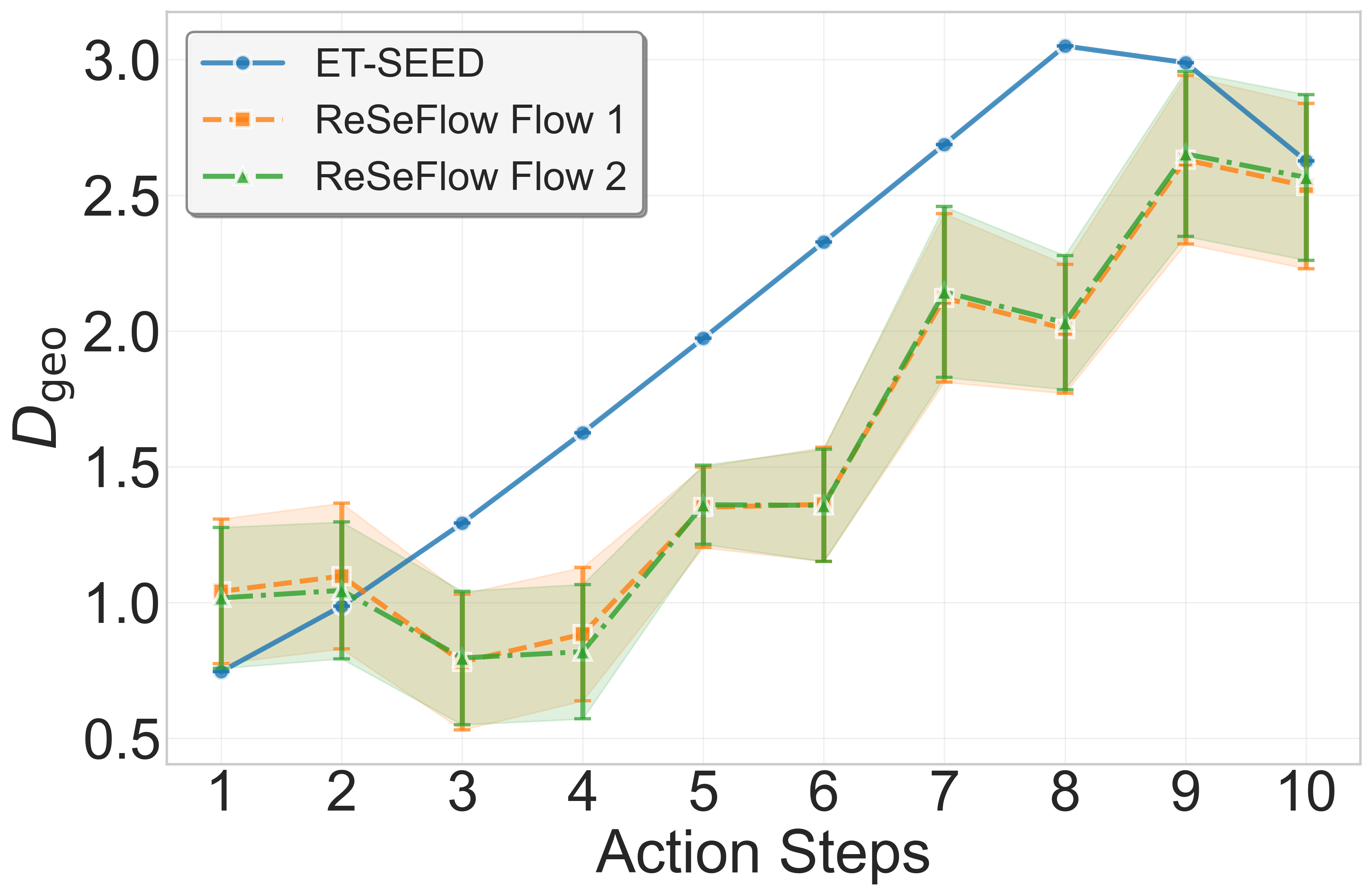}
    \caption{rf\_step = 2}
    \label{fig:tria-rf2}
  \end{subfigure}\hfill
  \begin{subfigure}{0.24\textwidth}
    \centering
    \includegraphics[width=\linewidth]{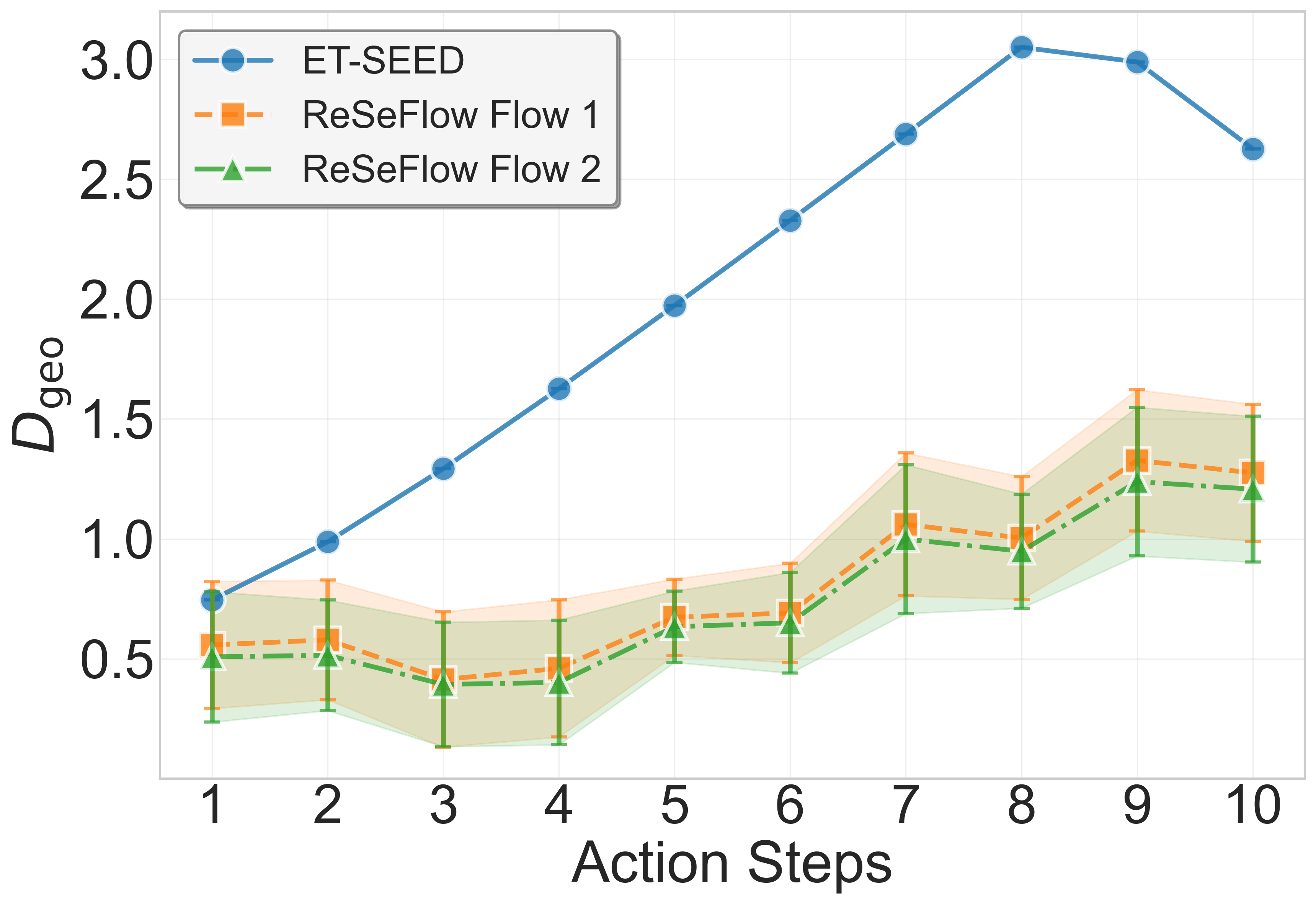}
    \caption{rf\_step = 50}
    \label{fig:tria-rf50}
  \end{subfigure}\hfill
  \begin{subfigure}{0.24\textwidth}
    \centering
    \includegraphics[width=\linewidth]{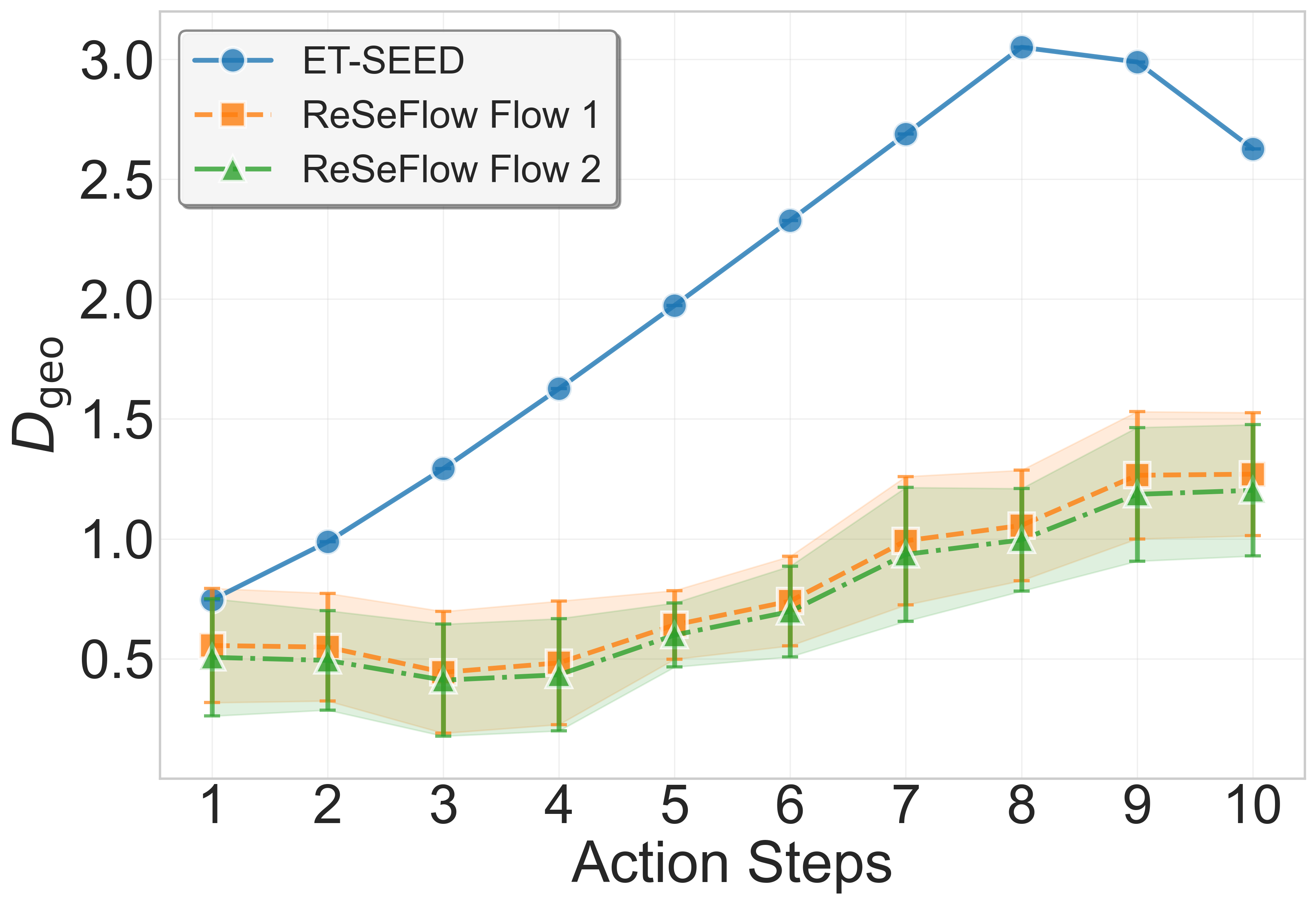}
    \caption{rf\_step = 100}
    \label{fig:tria-rf100}
  \end{subfigure}
  \caption{\textit{Rotating triangle} scene: distance over the full trajectory at ReSeFlow inference steps 1, 2, 50, and 100 (mean and variance over 10 seeds).}
  \label{fig:tria-trajectory-distance}
\end{figure*}

\subsection{Results}

We plot per-action $D_{\mathrm{geo}}$ curves and inference-step ablations for \textit{rotating triangle}, \textit{door opening}, and \textit{painting} from Fig.~\ref{fig:geodesic-compare} to Fig. \ref{fig:tria-trajectory-distance}. Each method is run with 10 random seeds, and solid lines show the mean and the shaded areas denote $\pm 1$ standard deviation (the lower the better). In detail, we compare the results in two groups as
\begin{itemize}
  \item \textbf{Fig. \ref{fig:geodesic-compare}}: $D_{\mathrm{geo}}$ versus the number of inference steps (rf\_step = 1, 2, 10, 50, and 100, respectively), revealing the step-performance trade-off of \ac{reseflow} and \ac{etseed}.
  \item \textbf{Fig.~\ref{fig:door-trajectory-distance}--\ref{fig:tria-trajectory-distance}}: per-action $D_{\mathrm{geo}}$ along the full trajectory (actions 1-10), comparing \ac{etseed} (with 100 inference steps) and \ac{reseflow}.
\end{itemize}
\setlength{\textfloatsep}{6pt plus 0pt minus 0pt}
\begin{table}[!t]
\caption{Comparison of the average $\overline{D}_{geo}$ for \ac{etseed} and Flow 1 and 2 for \ac{reseflow}.}
\label{tab:res_dis}
\begin{center}
\begin{tabular}{|c||c||c||c|}
\hline
Method  & \makecell{\textit{Door}\\\textit{Opening}} & \textit{Painting} & \makecell{\textit{Rotating}\\\textit{Triangle}}\\
\hline
\ac{etseed} & 2.360 & 1.756 & 2.087\\
\ac{reseflow} Flow 1 (Ours) & 0.487 & 0.767 & 0.959\\
\ac{reseflow} Flow 2 (Ours) & 0.450 & 0.766 & 0.876\\
\hline
\end{tabular}
\end{center}
\end{table}

We draw the distance figure of the whole trajectory generated for all scenes, and 10 seeds are selected to generate the mean and variance lines as in the figures mentioned above. Also, in Table \ref{tab:res_dis} with the average $\overline{D}_{geo}$, we evaluate \ac{reseflow} against the \ac{etseed} baseline across three challenging manipulation tasks. The results consistently demonstrate the superiority of our proposed framework in terms of both accuracy and inference efficiency.

\begin{figure}[h]
    \centering
    \includegraphics[width=0.5\textwidth]{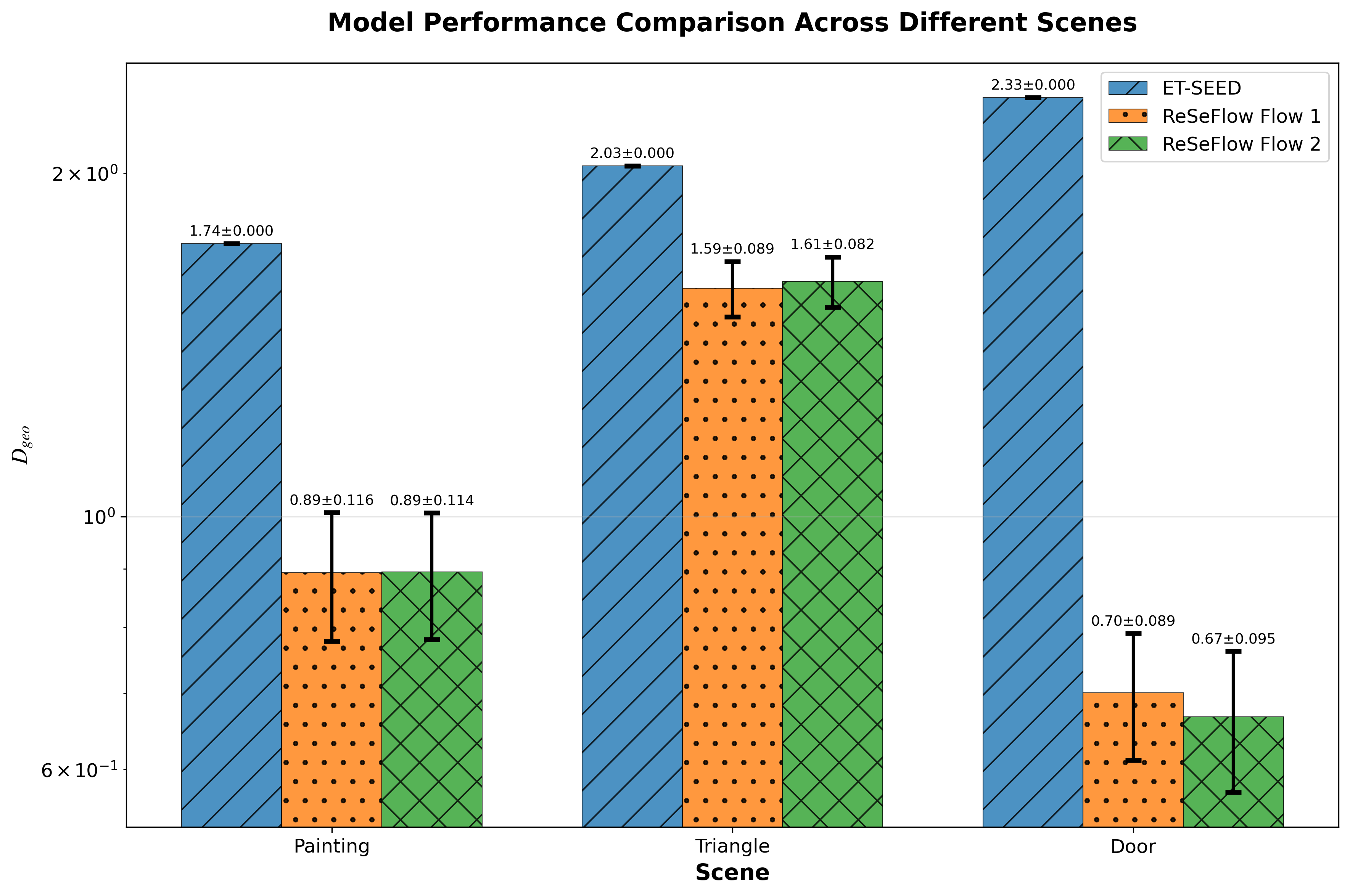}
    \caption{Model Performance Comparison Bar Chart Across Different Scenes. This figure compares the $D_{geo}$ performance of three models (\ac{etseed}, \ac{reseflow} Flow 1, and \ac{reseflow} Flow 2) across three scenes (\textit{painting}, \textit{rotating triangle}, and \textit{door opening}). Bar heights represent mean $D_{geo}$ values, while error bars show the standard deviation range across 10 runs. The $y$-axis uses logarithmic scale to better visualize performance differences across different orders of magnitude. Data is sourced from rf\_step=1 experimental results.}
    \label{fig:overall_performance_comparison}
\end{figure}

\textbf{Overall Performance}  As illustrated by the aggregate performance comparison in Fig. \ref{fig:overall_performance_comparison} and the radar chart in Fig.   \ref{fig:model_radar_chart}, both \ac{reseflow} Flow 1 and Flow 2 significantly outperform ET-SEED across all evaluated scenes. Across all three tasks, \ac{reseflow} consistently outperforms diffusion-based ET-SEED in most settings (Fig.   \ref{fig:overall_performance_comparison}): on \textit{rotating triangle}, \ac{reseflow} reaches about 1.59-1.61 at 1 step versus \ac{etseed} at $\sim$2.03 (100 steps), achieving a \textbf{21.9\% error reduction}; the gap widens on \textit{door opening} (Flow 2 drops from $\sim$0.67 to $\sim$0.36 with more steps, while \ac{etseed} stays around $\sim$2.33 (Fig. \ref{fig:geodesic-compare})); on \textit{painting}, \ac{reseflow} improves from $\sim$0.89 to $\sim$0.75 versus ET-SEED at $\sim$1.74, achieving a \textbf{48.5\% error reduction}.
\\

\textbf{Inference Efficiency} A key advantage of our method is its exceptional inference efficiency. Fig. \ref{fig:geodesic-compare} provides a detailed analysis of performance as a function of the number of inference steps. Notably, \ac{reseflow} achieves remarkable accuracy with only a single inference step. In the \textit{door opening} task (Fig.~\ref{fig:geodesic-compare}a), \ac{reseflow}'s single-step performance ($\sim$0.6--0.7) is vastly superior to ET-SEED's 100 inference-step performance ($\sim$2.3). While \ac{reseflow}'s accuracy marginally improves with additional steps (from 0.67 to 0.36 in the \textit{door opening} task), its one-shot generation capability represents a significant leap in computational efficiency, making it highly suitable for real-time applications.

\begin{figure}[!t]
    \centering
    \includegraphics[width=0.5\textwidth]{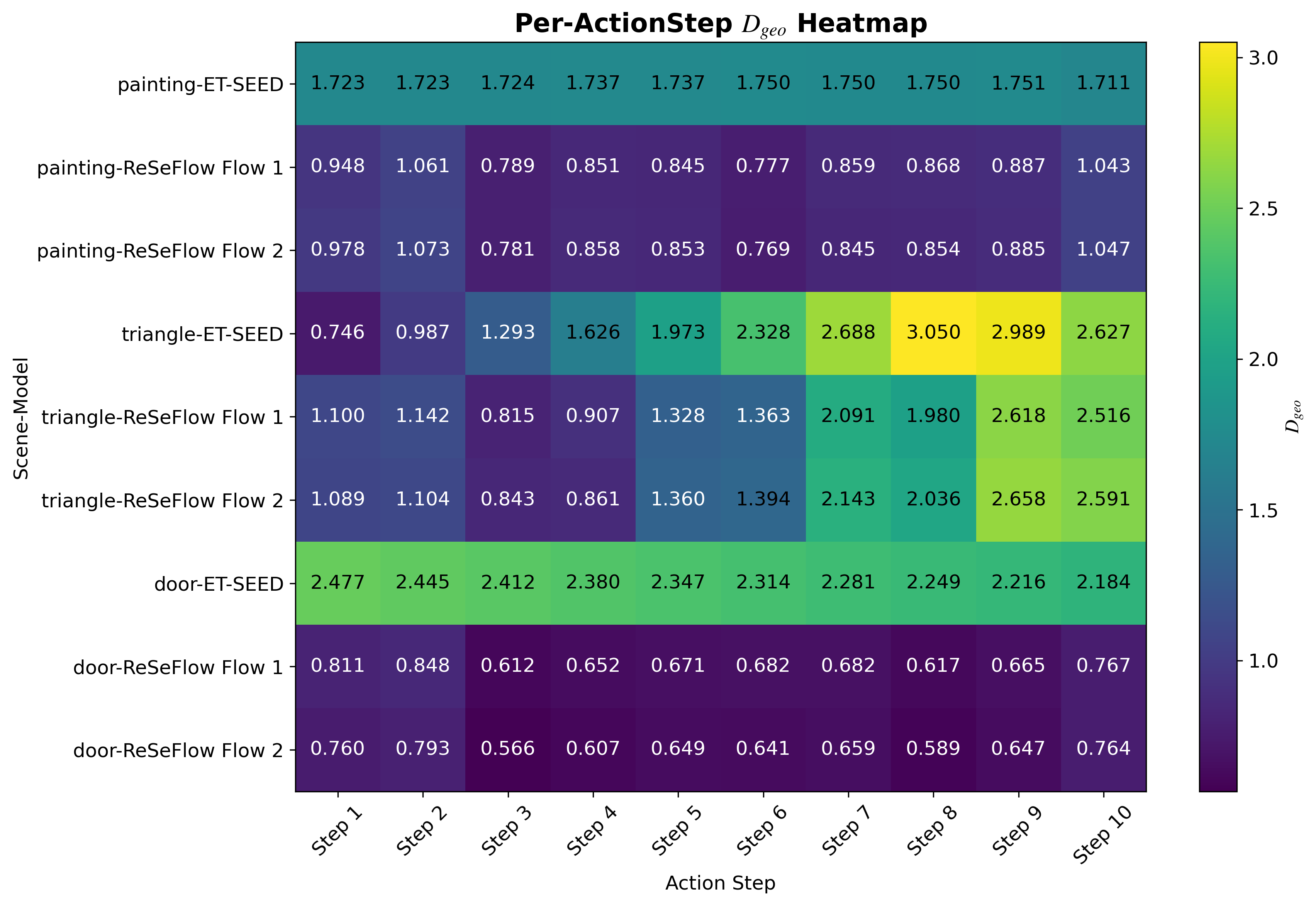}
    \caption{Per-ActionStep $D_{geo}$ Performance Heatmap. This figure illustrates the $D_{geo}$ performance variation of different scene-model combinations across individual action steps. Darker colors indicate higher $D_{geo}$ values (worse performance), with each cell displaying specific numerical values. The heatmap is based on per-action means data from rf\_step=1 experiments, reflecting the performance stability of models across different action steps.}
    \label{fig:per_action_heatmap}
\end{figure}

\textbf{Trajectory-Level Stability} We further analyze the per-action performance throughout the trajectory horizon. The heatmap in Fig. \ref{fig:per_action_heatmap} reveals that \ac{reseflow} maintains a consistently low $D_{\text{geo}}$ across all 10 action steps for all tasks, as indicated by the dark, uniform coloring of its corresponding rows. In contrast, \ac{etseed} exhibits higher and, in some cases (the \textit{rotating triangle} task), escalating error in later action steps. This is further corroborated by the trajectory plots in Fig.~\ref{fig:door-trajectory-distance}--\ref{fig:tria-trajectory-distance}, which show that \ac{reseflow} generates stable, low-error trajectories from the outset, independent of the number of inference steps. This highlights the model's ability to produce not only accurate final poses, but geometrically coherent and stable action sequences.


\subsection{Limitation and Discussion}

The proposed \ac{reseflow} achieves both data- and inference-efficient robotic policy generation, enabling the potential of areal-world application. The empirical results presented unequivocally highlight the advantages of the ReSeFlow framework. The core strength of our approach stems from the novel synergy between \ac{se3}-equivariant network architectures and the principles of rectified flow.

The exceptional single-step inference performance is directly attributable to the rectified flow component. By learning to transport policies along straight, geodesic-like paths in the model's latent space, \ac{reseflow} circumvents the need for the iterative, multi-step refinement process inherent to diffusion-based models like \ac{etseed}. This rectification process is the primary driver behind the dramatic reduction in computational cost without sacrificing accuracy, addressing a critical bottleneck for deploying generative policies in real-world scenarios.

Simultaneously, the enforcement of \ac{se3}-equivariance ensures that the model respects the underlying geometric symmetries of the manipulation tasks. This geometric prior provides strong inductive bias, leading to robust generalization from only a few demonstrations and consistent performance across varied object poses and configurations. The consistently low geodesic error across different scenes is a testament to this inherent robustness.

Furthermore, the performance of \ac{reseflow} Flow 2, which is trained by partially using a reflow procedure with data generated by Flow 1, demonstrates the stability and effectiveness of our two-stage training strategy. The results show that Flow 2 maintains the high level of accuracy established by Flow 1, confirming that the reflow process effectively refines the policy without introducing performance degradation. However, the \ac{reseflow} is only verified on the simulated data, the real-world case studies via our proposed method retain the main limitation.

\section{CONCLUSIONS}
\label{sec:conc}
In this paper, we introduced \ac{reseflow}, a novel framework for robotic policy learning that rectifies \ac{se3}-equivariant flows. By synergizing the geometric consistency of \ac{se3}-equivariance with the inference efficiency of rectified flows, ReSeFlow establishes a new paradigm for generating robust and long-horizon manipulation trajectories. Our experiments demonstrate that it decisively outperforms existing diffusion-based methods, achieving superior accuracy with significantly fewer inference steps. In summary, by combining the efficiency of rectified flows with the geometric robustness of \ac{se3}-equivariance, ReSeFlow sets a new technical benchmark for data-efficient and inference-efficient robotic policy learning. We believe this work presents a promising direction for deploying complex generative policies in real-world robotic systems. Future work will focus on adaptive step sizing and broader real-robot validations to further demonstrate its universality and transferability.








\bibliographystyle{IEEEtran}
\bibliography{reference}
\end{document}